\documentclass[lettersize,journal]{IEEEtran}
\usepackage{amsmath,amsfonts}
\usepackage{algorithmic}
\usepackage{algorithm}
\usepackage{array}
\usepackage[caption=false,font=normalsize,labelfont=sf,textfont=sf]{subfig}
\usepackage{textcomp}
\usepackage{xcolor}
\usepackage{stfloats}
\usepackage{url}
\usepackage{verbatim}
\usepackage{graphicx}
\usepackage{cite}
\usepackage{multirow}
\usepackage{xcolor}
\begin{document}


\title{Transcending Fusion: A Multi-Scale Alignment Method for Remote Sensing Image-Text Retrieval}


 \author{Rui Yang, Shuang Wang~\IEEEmembership{Member,~IEEE}, Yingping Han, Yuanheng Li, Dong Zhao, Dou Quan~\IEEEmembership{Member,~IEEE}, Yanhe Guo~\IEEEmembership{Member,~IEEE}, and Licheng Jiao~\IEEEmembership{Fellow,~IEEE}

\thanks{Rui Yang, Shuang Wang, Yingping Han, Yuanheng Li, Dong Zhao, Dou Quan, Yanhe Guo, and Licheng Jiao are with the School of Artificial Intelligence, Xidian University, Xi'an 710071, China, also with the Key Laboratory of Intelligent Perception and Image Understanding of Ministry of Education, Xidian University, Xi’an 710071, China. (e-mail: shwang@mail.xidian.edu.cn).}}

\markboth{Journal of \LaTeX\ Class Files,~Vol.~14, No.~8, August~2021}
{Shell \MakeLowercase{\textit{et al.}}: A Sample Article Using IEEEtran.cls for IEEE Journals}


\maketitle

\begin{abstract}

Remote Sensing Image-Text Retrieval (RSITR) is pivotal for knowledge services and data mining in the remote sensing (RS) domain. Considering the multi-scale representations in image content and text vocabulary can enable the models to learn richer representations and enhance retrieval. Current multi-scale RSITR approaches typically align multi-scale fused image features with text features, but overlook aligning image-text pairs at distinct scales separately. This oversight restricts their ability to learn joint representations suitable for effective retrieval. We introduce a novel Multi-Scale Alignment (MSA) method to overcome this limitation. Our method comprises three key innovations: (1) Multi-scale Cross-Modal Alignment Transformer (MSCMAT), which computes cross-attention between single-scale image features and localized text features, integrating global textual context to derive a matching score matrix within a mini-batch, (2) a multi-scale cross-modal semantic alignment loss that enforces semantic alignment across scales, and (3) a cross-scale multi-modal semantic consistency loss that uses the matching matrix from the largest scale to guide alignment at smaller scales. We evaluated our method across multiple datasets, demonstrating its efficacy with various visual backbones and establishing its superiority over existing state-of-the-art methods. The GitHub URL for our project is: \textit{https://github.com/yr666666/MSA}
\end{abstract}



\section{Introduction}
\IEEEPARstart{W}{ith} the recent development of satellite and remote sensing (RS) technology, RS satellites produce massive amounts of RS images every day \cite{Akiva_2022_CVPR}. Retrieving valuable data from such massive RS image databases based on semantics is of great significance for RS data mining and knowledge services \cite{yang2022multimodal,yuan2021exploring}. To meet the increasing demand for RS image data search and mining, scholars are dedicated to researching remote sensing image-text retrieval (RSITR), which enables the mutual retrieval of RS images and text \cite{cheng2021deep,yuan2021exploring,yuan2022remote,yuan2021lightweight,tang2023interacting,zhang2023hypersphere,yuan2023parameter,10181233,pan2023prior}. 
 For example, Cheng et al. \cite{cheng2021deep} constructed a semantic alignment module to explore the relationship between optical RS images and corresponding text descriptions. Tang et al. \cite{tang2023interacting} proposed an interacting-enhancing feature transformer to mine deep links between RS images and text, achieving cross-modal matching.

\begin{figure*}[t]
\centering
\includegraphics[width=1\textwidth]{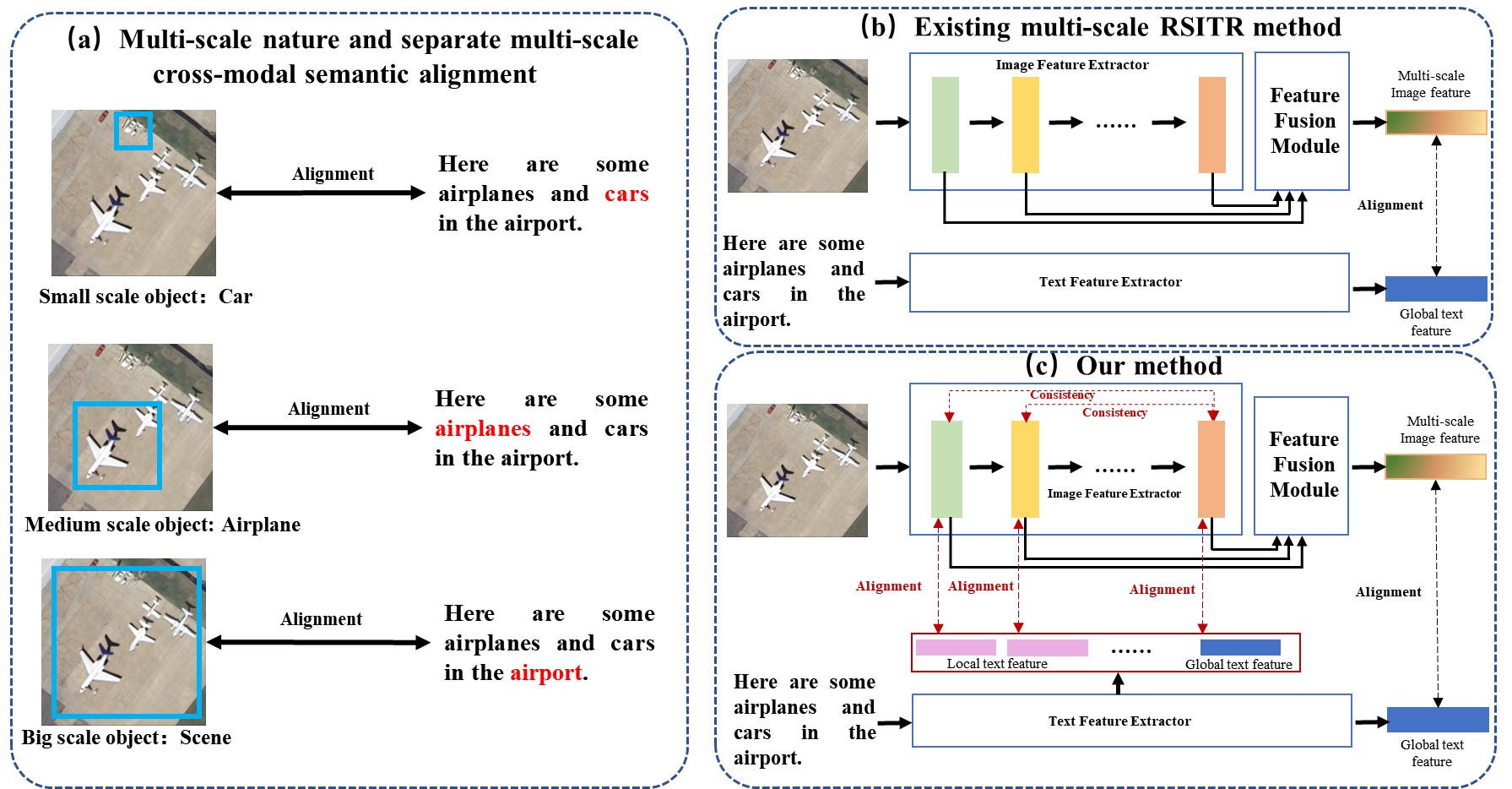} 
 \caption{Illustration of multi-scale and existing methods. (a) The multi-scale characteristics in RSITR and the separate alignment of image-text at different scales. (b) Existing RSITR methods based on multi-scale image fusion. (c) The method proposed in this paper to achieve separate alignment of image-text at different scales.}
\label{intro}
\end{figure*}

In the study of RSITR, we observe the presence of multi-scale representations in both image content and text vocabulary. As illustrated in Figure \ref{intro} (a), RS images contain targets that exhibit properties at various scales, such as cars, airplanes, and airports. Similarly, at the textual level, there are descriptions at different scales, such as "car," "some airplane," and "airport." These different scales in images correspond to varying levels of granularity in text descriptions. The global features of the RS image may align with the vocabulary term "airport" that describes the entire scene. Medium-scale RS image features may correspond to words like "some airplanes," while small-scale RS image features may align with the word "car." This phenomenon indicates the multi-scale nature of target representation in RS image (text) and the separate image-text alignments at multiple scales in RSITR.

Some researchers have proposed approaches to address the multi-scale nature of target representation in RS images \cite{yuan2021exploring, yuan2022remote,pan2023reducing,yuan2021lightweight}. As illustrated in Figure \ref{intro} (b), they handle the image and text through two branches. In the image branch, multiple scales of image features are extracted and combined to obtain multi-scale fused image features. In the text branch, text features are extracted using a text encoder. Subsequently, a global alignment is performed between the multi-scale fused image features and text features to achieve cross-modal retrieval. For instance, Yuan et al. \cite{yuan2021exploring} proposed a multi-scale visual self-attention module to obtain the multi-scale fused image feature used in retrieval. Pan et al. \cite{pan2023reducing} designed a method that combines multi-scale image feature fusion and text coarse-grained enhancement. These aforementioned RSITR methods \cite{yuan2021exploring,yuan2022remote, yuan2021lightweight,pan2023reducing} have demonstrated promising results. However, a significant unresolved issue remains. While they address the multi-scale nature of target representation in RS image, they overlook the separate image-text alignments at multiple scales in RSITR. Aligning only the multi-scale fused image features with text fails to account for this varying alignments, making it challenging to obtain a joint image-text representation that is better suited for retrieval. Consequently, suboptimal results are obtained.

To address this issue, this paper proposes considering separate image-text alignments at multiple scales in RSITR, as depicted in Figure \ref{intro} (c). Aligning image features with text features at different scales enhances \textbf{multi-scale cross-modal semantic alignment} in the joint representation of image and text, achieving more precise alignment beyond solely fusion methods. This optimization of RS image-text representation improves retrieval performance.
Furthermore, during separate alignment, varying levels of difficulties arise in aligning image with text at different scales (for a more detailed analysis, please refer to Section~\ref{Analysis of Cross-scale Multi-modal Consistency.}), as shown in Figure~\ref{csmmc}. This phenomenon indicates that image-text alignments at a smaller scale are relatively challenging. To overcome this issue and further enhance alignments at multiple scales, utilizing multimodal semantics from largest-scale image-text alignment as a supervisory signal can guide alignments of image and text at smaller scales. This improves \textbf{cross-scale multi-modal semantic consistency} of image-text alignments at multiple scales, enriching the joint representation across scales and further enhancing RSITR performance.

Based on the above discussion, this paper proposes a novel Multi-Scale Alignment method (MSA) for RSITR, which enables separate image-text alignments at different scales. The proposed method has three main innovations: (1) Multi-Scale Cross-Modal Alignment Transformer (MSCMAT), (2) Multi-Scale Cross-Modal semantic Alignment loss (MSCMA loss), and (3) Cross-Scale Multi-Modal semantic Consistency loss (CSMMC loss).
The MSCMAT, which is based on the Transformer \cite{vaswani2017attention} structure, utilizes the single-scale image feature as the query vector Q and the local text feature vectors as the key vector K and value vector V. By employing the cross-attention mechanism, MSCMAT can adaptively learn the alignment between the single-scale image feature and text information, while simultaneously incorporating global textual information and outputting a matching score matrix within a mini-batch.
The MSCMA loss constrains the image-text matching score matrix outputted by MSCMAT between at each single-scale through the use of contrastive loss, ensuring the multi-scale cross-modal semantic alignment.
The CSMMC loss leverages the image-text matching score matrix between the largest-scale image feature and text as a teacher signal. It aligns the distribution of the matching score matrix between the largest-scale image and text with that of the other-scale image-text matching score matrix, ensuring cross-scale multi-modal semantic consistency. Using the aforementioned three components, we can achieve separate image-text alignments at different scales.

After separately aligning image-text at different scales, a selective attention mechanism based on the SENet \cite{hu2018squeeze} is employed to adaptively fuse the image features from multiple scales, resulting in the final multi-scale fused image feature. The multi-scale fused image feature and text feature are further optimized using a triplet loss, enabling cross-modal matching in a shared space. During testing, only the multi-scale fused image feature is employed for retrieval, maintaining a concise dual-stream retrieval architecture to ensure retrieval efficiency.

The proposed MSA method is systematically evaluated on three widely used optical RS image-text retrieval datasets. The experimental results demonstrate the effectiveness of the proposed method and achieve state-of-the-art retrieval performance.

Overall, the contributions of this paper are as follows.

\begin{itemize}
 \item{We propose a novel MSA method for RSITR, which surpasses existing methods by considering image-text alignment at different scales. The MSA goes beyond relying solely on multi-scale fused image features and allows the model to learn intricate alignment relationships and obtain richer feature representations.}

\item{We designed an MSCMAT, which utilizes the cross-attention mechanism to obtain alignment scores between RS image and text at different scales.}

\item{We proposed an MSCMA loss that enforces image-text alignment at different scales by constraining their cross-modal semantic alignment.}

\item{We proposed a CSMMC loss that bridges the difficulty gap in aligning image-text at different scales by enforcing cross-scale multi-modal semantic consistency.}

 \end{itemize}

In Section II, we introduce the related work of this paper, and in Section III, we describe the proposed method. Section IV depicts the experiments. In Section V, we conclude the work of this paper.

\section{Related Works}

In this section, we first introduce natural cross-modal image-text retrieval, followed by RS cross-modal image-text retrieval. Lastly, we provide an overview of multi-scale methods in RS image processing.

\subsection{Natural image-text retrieval}
Natural image-text retrieval is a data retrieval technique that searches for the most matching or relevant data samples from a database of one modality (text or image) based on a query of another modality (image or text) \cite{lee2018stacked}. The natural image-text retrieval  methods can be divided into dual-flow \cite{Karpathy_2015_CVPR,Huang_2018_CVPR,Chen_2020_CVPR,lee2018stacked,chun2021probabilistic,yang2023knowledge,yang2024continual} and single-flow \cite{wang2019camp,lu2019vilbert,chen2020uniter,huang2020pixel,li2020oscar,kim2021vilt,yuan2021florence} methods according to the data flow. The dual-flow models are less interactive but fast, while the single-flow models are highly interactive but slow.

\subsubsection{Dual-flow Methods}
Dual-flow methods \cite{Karpathy_2015_CVPR,Huang_2018_CVPR,Li_2019_ICCV,chun2021probabilistic} use two encoders to extract features from images and text, respectively. They then calculate the distance between the features by inner product or cosine similarity and sort the samples based on the distance to achieve nearest neighbor retrieval.
Zhen et al. \cite{zhen2019deep} proposed a cross-modal retrieval method called deep supervised cross-modal retrieval, which utilizes category supervised information to find a common representation space. Li et al. \cite{Li_2019_ICCV} proposed a graph convolution-based image representation generation model to generate visual representations of critical objects and semantic concepts that capture the scene. Chun et al. \cite{chun2021probabilistic} suggested constructing a probabilistic cross-modal embedding space, in which samples from different modalities are represented as probability distributions.

Dual-flow methods have the advantage of performing offline feature extraction and indexing in advance, which makes online retrieval faster. However, the use of separate encoders for the two modalities limits retrieval accuracy due to the lack of inter-modal interaction. Despite this drawback, the dual-flow model is still necessary for efficient retrieval in large databases.

\subsubsection{Single-flow Methods}

In contrast, single-flow methods \cite{lu2019vilbert,chen2020uniter,huang2020pixel,li2020oscar,kim2021vilt,yuan2021florence} use the Transformer module \cite{vaswani2017attention} to construct a visual-language pretraining (VLP) model that calculates cross-attention between images and text to achieve cross-modal interactions. The model outputs matching scores for nearest neighbor retrieval. 
Kim et al. \cite{kim2021vilt} proposed a convolution-free method, VILT, where image patches and text words are jointly entered as tokens into the Transformer encoder to output matching scores. Lu et al. \cite{yuan2021florence} proposed a multimodal learning method, Florence, which uses Roberta \cite{liu2019roberta} as a text encoder and co-swin \cite{liu2021swin} as an image encoder to compute the cross-attention of text and image.

Single-flow methods improve retrieval accuracy by enabling cross-modal interaction, but this significantly increases the time cost at the test stage. It is impractical to use a single-flow method for retrieving images with vast amounts of data in RS scenes. 

In order to maintain the retrieval efficiency of single-flow methods and the multimodal interactivity of dual-flow methods, the proposed MSA adopts a single-flow structure during training. However, during testing, the MSCMAT does not participate in the forward propagation, resulting in a dual-flow structure of the model.

\subsection{Remote sensing image-text retrieval (RSITR)}

With the demand for automatic mining of RS images, more and more studies are focusing on RSITR. RSITR can play an essential role in RS data services and semantic localization tasks \cite{yuan2021exploring}, so it has received more attention in recent years and has become a research hotspot. Cheng et al. \cite{cheng2021deep} introduced a cross-modal image-text retrieval network by using attentional and gating mechanisms to establish a direct relationship between RS images and paired text data. Yuan et al. \cite{yuan2021exploring} proposed an asymmetric multi-modal feature matching network for RSITR, using multi-scale image features to guide the representation of text features to achieve image-text interaction. Yuan et al. \cite{yuan2022remote} proposed a framework based on global and local information and designed a multi-level information dynamic fusion module to efficiently integrate different levels of image features. Recently, Yuan et al. \cite{yuan2021lightweight} designed a lightweight cross-modal RS image retrieval network to reduce inference time. Rahhal et al. \cite{9925582} designed transformer-based RS image and text encoders for multilingual RSITR. Yu et al. \cite{9999008} proposed a RS image-text matching framework based on a graph neural network to solve the problem of asymmetric text and image information.

\subsection{Multi-scale methods in RS image processing}

Existing methods for multi-scale RS image processing primarily focus on the fusion of multi-scale RS image features. After obtaining the fused features, downstream tasks such as semantic segmentation, object detection, and scene classification can be performed. Based on the data flow of fused features, existing methods can be divided into two categories: multi-level features fusion-based methods and multi-scale inputs fusion-based methods.

\subsubsection{Multi-level features fusion-based methods}
In multi-level features fusion-based methods \cite{peng2019densely,fu2020rotation,tang2022emtcal,chen2023efcomff,xiao2023enhancing,zheng2023remote,hou2022attention,zhou2023misnet}, the input images are fed into a convolutional neural network, and the convolutional image features from different levels (receptive fields) are fused to obtain multi-scale fused features. 
Tang et al. \cite{tang2022emtcal} designed a multi-scale Transformer and cross-level attention mechanism to aggregate features from different levels of a convolutional neural network for RS scene classification. Chen et al. \cite{chen2023efcomff} proposed a feature correlation enhancement module to enhance the correlation between features from different layers of a CNN network. They then fused the enhanced multi-level features to obtain fusion features for RS scene classification.

\subsubsection{Multi-scale inputs fusion-based methods}
These methods \cite{wang2020looking, yin2022high, qian2023multi} involve cropping and scaling the RS images to obtain inputs of different scales. These images are then fed into feature extractors to obtain corresponding image features. The fusion of these features results in multi-scale fused features. 
Yin et al. \cite{yin2022high} performed cropping and scaling on RS images to create three scales: large, medium, and local. These scales were input into a feature extractor, and semantic segmentation was achieved through fusion based on position attention and channel attention mechanisms.
Qian \cite{qian2023multi} introduced the multi-scale image splitting-based feature enhancement framework, which involves image cropping into multiple scales. By utilizing fusion methods based on spatial attention, the framework generates fused features for weakly supervised RS object detection.

Researchers have applied feature fusion-based multi-scale methods to RSITR \cite{yuan2021exploring,yuan2022remote,zheng2023fine,pan2023reducing,chen2023multiscale}. Pan et al. \cite{pan2023reducing} developed a method integrating multi-scale image feature fusion with coarse-grained text enhancement, while Yuan et al. \cite{yuan2022remote} introduced a network for cross-modal RS image retrieval that merges global and local information. A closely related work, FAAMI \cite{zheng2023fine}, introduces a fine-grained semantic alignment method addressing the complex interactions between multi-scale image and textual features. Our approach significantly diverges from FAAMI and other methods in targeting specific problems, methodological design, and data flow, particularly in addressing the alignment of image-text pairs across different scales, which previous methods have overlooked. In contrast to previous multi-scale processing methods, this paper introduces a novel MSA method specifically for RSITR. Table~\ref{tab:0} highlights the differences between our method and other multi-scale RSITR approaches.

\begin{table}[]

\caption{\label{tab:0}Comparison of MSA with various multi-scale RSITR methods}
\resizebox{\linewidth}{!}{
\begin{tabular}{ccccccc}
\hline
\textit{Method}  & \textit{\begin{tabular}[c]{@{}c@{}}Multi-scale \\ fusion\end{tabular}} & \textit{\begin{tabular}[c]{@{}c@{}}Alignment at \\ different scales\end{tabular}} & \textit{\begin{tabular}[c]{@{}c@{}}Single/\\ Dual-flow\end{tabular}} & \textit{\begin{tabular}[c]{@{}c@{}}Image \\ backbone\end{tabular}} & \textit{\begin{tabular}[c]{@{}c@{}}Text \\ backbone\end{tabular}} & \textit{mR on RSITMD} \\ \hline
AMFMN \textit{TGRS'22}    & Yes                                                                    & No                                                                                          & Single                                                                       & ResNet                                                            & Bi-GRU                                                            & 29.72                 \\
SWAN \textit{ICMR'23}    & Yes                                                                    & No                                                                                           & Dual                                                                        & ResNet                                                             & Bi-GRU                                                           & 34.11                \\
FAAMI \textit{Sensors'23} & Yes                                                                   & No                                                                                          &Single                                                                        &DetNet                                                             & BERT                                                              & 35.99                \\
GaLR \textit{TGRS'22}     & Yes                                                                   & No                                                                                           & Dual                                                                          & GCN, ResNet                                                       & Bi-GRU                                                             & 31.41                \\
MSITA \textit{TGRS'24}    & Yes                                                                    & No                                                                                           & Single                                                                       & Transformer                                                        & Bi-GRU                                                            & 34.48                \\
MSA             & Yes                                                                    & Yes                                                                                          & Dual                                                                          & ResNet                                                             & BERT                                                              & \textbf{38.08}                 \\ \hline
\end{tabular}}
\end{table}

\begin{figure*}[t]
\centering
\includegraphics[width=1\textwidth]{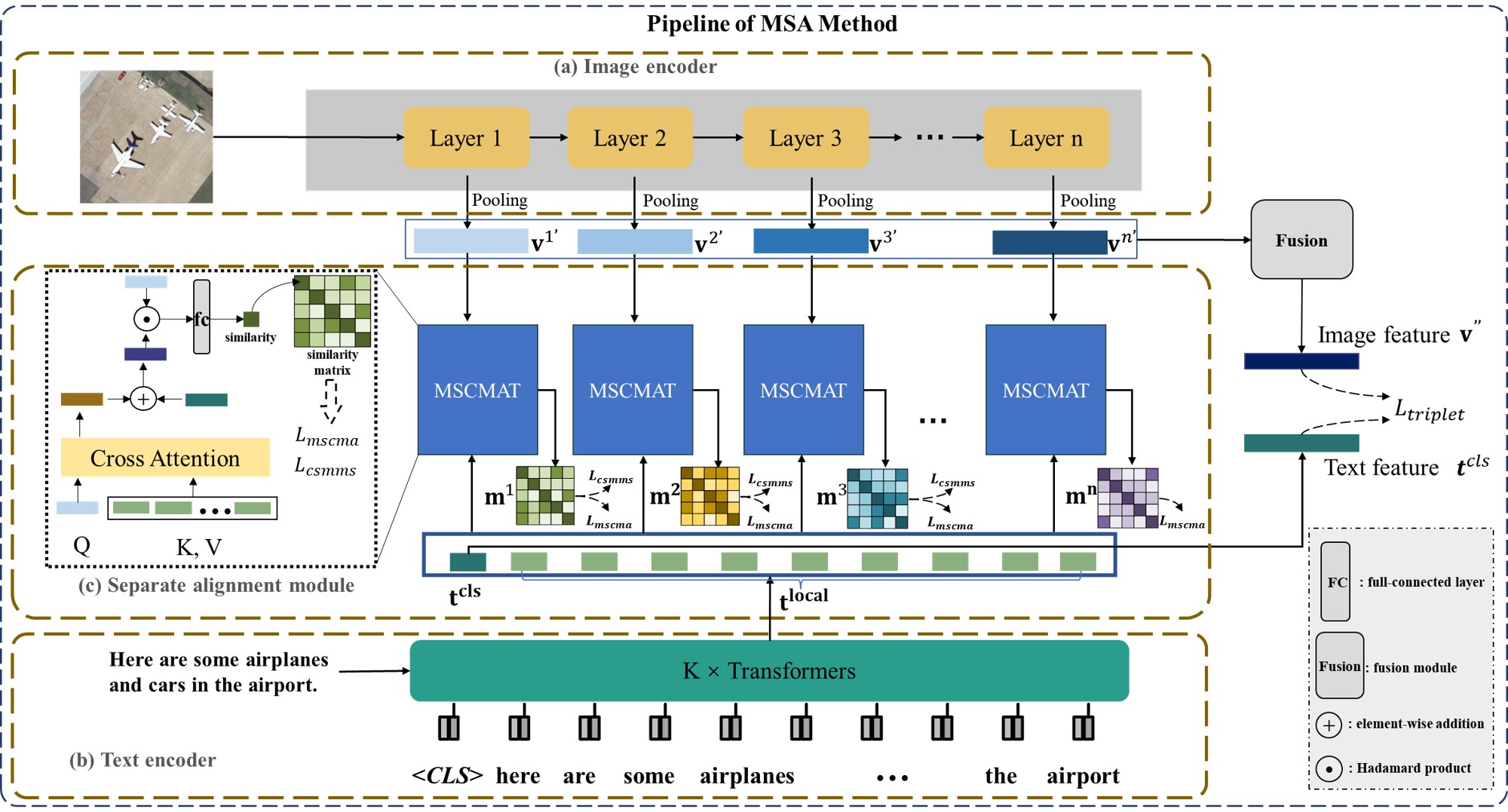} 
 \caption{The pipeline of MSA consists of three parts. (a) The multi-scale image feature extractor, where this paper utilizes the ResNet (ResNet-18, ResNet-50, ResNet-101). (b) The text feature extractor, where BERT is employed to extract both local text features and the CLS global feature. (c) The innovation of this paper, which includes MSCMAT, MSCMA loss, and CMMSC loss. MSCMAT adaptively learns the image-text alignment at different scales between the image features and text features. The two proposed losses respectively enhance the cross-modal semantic alignment and cross-modal semantic consistency of the model.}
\label{pipeline}
\end{figure*}

\section{Method}
To enhance retrieval performance by achieving separate image-text alignments at different scales, this paper proposes a novel MSA method for RSITR. The MSA method has three key innovations: MSCMAT, MSCMA loss, and CSMMC loss. In the following sections, we will first provide an overview of the method and then delve into detailed explanations of the three key innovations. 

\subsection{Overall Method}

The structure of MSA consists of four parts: an image encoder, a text encoder, a separate alignment module, and a fusion module, as shown in Figure~\ref{pipeline}. The three innovations of this paper are within the separate alignment module.

The image encoder (such as ResNet \cite{he2016deep}) is used to extract image features at multiple scales, while the text encoder (such as BERT \cite{devlin2018bert}) is used to extract text features. The obtained image features and text features at each scale are separately inputted into an MSCMAT, resulting in image-text alignment score matrices within a mini-batch at different scales. The alignment score matrices are then optimized using the MSCMA loss and the CSMMC loss, enhancing the multi-scale cross-modal semantic alignment and cross-scale multi-modal semantic consistency of the multi-scale image-text joint representations.
Simultaneously, the image features at multiple scales are passed through a fusion module based on a selective attention mechanism \cite{hu2018squeeze}, resulting in multi-scale fused image features. The multi-scale fused image features and text features are optimized using a triplet loss, which brings matching image-text pairs closer and non-matching pairs further apart, thereby achieving image-text matching.

We will explain the data flow using a batch of image-text pairs as input, with a batch size of $b$.
Let $D = \{(I_i,T_i)\}$, $i\in(1,N)$ represent a training dataset consisting of $N$ image-text pairs. The proposed method utilizes the dataset $D$ to train a retrieval model, which can be used for RSITR on the test set.

Firstly, we introduce the image encoder. The image encoder can be represented as follows:
\begin{equation} 
\label{eq1}
\mathbf{v}^1, \mathbf{v}^2, \ldots, \mathbf{v}^n = \text{Pool}(\text{Resnet}(\mathbf{I}))
\end{equation}
Here, $\textbf{I}$ represents a batch of input images, and $\mathbf{v}^i$ represents the single-scale features obtained from the $i-th$ convolutional layer of ResNet (such as \textit{conv\_2\_x}, \textit{conv\_3\_x}, \textit{conv\_4\_x}, \textit{conv\_5\_x}) after spatial pooling. The term $Pool$ refers to the spatial pooling operation. In theory, we can use various visual models as multi-scale image encoder. In practice, we use the ResNet series models (ResNet18, ResNet50, and ResNet101).

After obtaining the image features at multiple scales, we use linear layers to map the features from different scales to the same dimensionality, enabling alignment with the text in the subsequent MSCMAT, shown as in Equation.~\ref{eq2}.

\begin{equation}
\label{eq2}
\mathbf{v}^{i\prime}=L^i\left(\mathbf{v}^i\right)
\end{equation}
$\mathbf{v}^{i\prime} $ represents the tensors of image features at different scales within a batch, with dimensions of $b \times d$. Here, $b$ denotes the batch size and $d$ represents the feature dimensionality. $L^i$ refers to the fully connected layer. $i$ represents the i-th scale.

We utilize the text encoder BERT to extract text features.
\begin{equation}
\mathbf{t}^{cls},\mathbf{t}^{local}=BERT(\mathbf{T}))
\label{eq3}
\end{equation}
Here, $\mathbf{t}^{cls}\in\mathbb{R}^{b\times d}$ represents the output corresponding to the cls-token of BERT, $\mathbf{t}^{local}=\{\mathbf{t}_{1}^{local},\mathbf{t}_{2}^{local},\ldots,\mathbf{t}_{p}^{local}\},\mathbf{t}_{i}^{local}\in\mathbb{R}^{{b}\times{p}\times{d}}$ represents the output feature corresponding to the input word, and $p$ represents the length of the text.

Then, we discuss the separate alignment module.
We input the image features and text features at different scales, from a batch of image-text pairs, into the MSCMT (which will be explained in detail in the next section), generating image-text similarity matrices at different scales, as shown in Equation~\ref{eq4}.
\begin{equation}
\mathbf{m}^{i}=MSCMT\left(\mathbf{v}^{{i}\prime},\mathbf{t}^{cls},\mathbf{t}^{{local}}\right)
\label{eq4}
\end{equation}
Here, $\mathbf{m}^{i}$ represents the image-text similarity matrix at the i-th scale, with dimensions of ${b}\times{b}$.

Next, we optimize the cross-modal similarity matrices at different scales and text using the MSCMA loss ($L_{MSCMA}$) and the CSMMC loss ($L_{CSMMC}$) (which will be explained in detail in the subsequent sections).

To enable image-text retrieval, we fuse the image features from multiple scales to obtain fused image features, as shown in Equation~\ref{eq5} and Equation~\ref{eq6}.

\begin{equation}
\mathbf{v}^{\prime\prime}=\mathbf{A}^{ms}\odot\left(\mathbf{v}^{{1}\prime}\oplus\mathbf{v}^{{2}\prime}\oplus\ldots\oplus\mathbf{v}^{{n}\prime}\right)
\label{eq5}
\end{equation}

\begin{equation}
\mathbf{A}^{ms}=Sigmoid(L^{ms}\left(\mathbf{v}^{{1}\prime}\oplus\mathbf{v}^{{2}\prime}\oplus\ldots\oplus\mathbf{v}^{{n}\prime}\right))
\label{eq6}
\end{equation}
Here, $L^{ms}$ represents a fully connected layer, $\odot$ denotes element-wise multiplication, $\oplus$ denotes element-wise addition, and $\mathbf{v}^{\prime\prime} \in \mathbb{R}^{b\times d}$ represents the multi-scale fused image features within a batch.

We optimize the multi-scale fused image features and global text features using the triplet loss. This loss function aims to minimize the distance between matching image-text pairs in a shared space while maximizing the distance between non-matching pairs, as shown in Equation.~\ref{eq7}.

\begin{equation}
\begin{split}
L_{tri}=\sum_{i=1}^{b}{Max(0,d-\ S({\mathbf{v}^{\prime\prime}}_{{i}\ },{\mathbf{t}^{{cls}}}_j^+)\ +\ S({\mathbf{v}^{\prime\prime}}_{{i}\ },{\mathbf{t}^{{cls}}}_k^-))}+ \\
\sum_{i=1}^{b}{Max(0,d-\ S(\mathbf{t}_{i}^{{cls}},{v^{\prime\prime}}_j^+)\ +\ S(\mathbf{t}_{i}^{{cls}},{v^{\prime\prime}}_k^+))}
\end{split}
\label{eq7}
\end{equation}
Here, $b$ represents the batch size, $j$ represents the index of the positive sample, $k$ represents the index of the negative sample, $S()$ denotes cosine similarity, and $d$ represents the margin.

We perform end-to-end training on the proposed method. The overall loss is defined as follows:
\begin{equation}
L_{total}=L_{tri}+\alpha\ {\ast\ L}_{MSCMA}+\beta\ \ast{\ L}_{CSMMC\ }
\label{eq8}
\end{equation}
Here $\alpha$ and $\beta$ are combination coefficients, used to weight the individual loss terms in the total loss function. 

Subsequently, we will provide a detailed explanation of the three components included in the proposed method: MSCMAT, $L_{MSCMA}$, and $L_{CSMMC}$.

\subsection{Multi-Scale Cross-Modal Alignment Transformer (MSCMAT)}

The MSCMAT utilizes a multi-head cross-attention to compute the image-text alignment score at each single scale. 

The individual MSCMAT used in the Equation~\ref{eq4} maintains the same structure. It takes as input the image features at a single scale and text features (including the $cls$ feature and local features) and outputs the alignment matrix between the image and text at that specific scale. We treat the image features at a single scale as query $Q$ and the local text features as key $K$ and value $V$. The cross-attention mechanism produces an aggregated text vector with the same dimensions as $Q$. This vector is a weighted aggregation of the local text vectors based on the similarity between the single-scale image feature and the local text vectors, representing the alignment and aggregation of individual vocabulary words with the single-scale image feature. This aggregated text vector can adaptively achieve fine-grained alignment between the single-scale image feature and the text. The equations below illustrate this process.

\begin{equation}
Q\ =\ \mathbf{v}^{{i}\prime}{\ \mathbf{W}}^{q}
\label{eq9}
\end{equation}

\begin{equation}
K\ =\ \mathbf{t}^{{local}\prime}{\ \mathbf{W}}^{k}
\label{eq10}
\end{equation}

\begin{equation}
V\ =\ \mathbf{t}^{{local}\prime}{\ \mathbf{W}}^{v}
\label{eq11}
\end{equation}

\begin{equation}
\mathbf{t}^{{agg}}=Att(Q,K,V)=\ Sigmod(\frac{QV^T}{\tau})\ V
\label{eq12}
\end{equation}
$\mathbf{t}^{{agg}}\in\mathbb{R}^{m\times d}$ represents the aggregated text vector. We employ a multi-head attention mechanism, with the number of heads set to 8. $\tau$ denotes the scaling factor.

After obtaining the aggregated text vector, we add it to the $cls$ feature to obtain a text feature vector that incorporates global information. Then, we compute the similarity between the text feature vector with global information and the image feature vector at a single scale. This similarity is considered as the image-text alignment score at the single-scale, as shown in Equation~\ref{eq13}

\begin{equation}
\mathbf{m}^{i}= \mathbf{v}^{{i}\prime} {(\mathbf{t}^{{agg}} \oplus \mathbf{t}^{cls})}
\label{eq13}
\end{equation}

In the specific implementation of this paper, we set the number of image scales to 4 and the number of attention layers in each MSCMAT to 1. Therefore, all the MSCMATs incorporate four multi-head cross-attention layers.

\subsection{ Multi-Scale Cross-Modal semantic Alignment loss (MSCMA loss)}
The essence of retrieval tasks is to achieve semantic alignment between image features and text features, while RS images-text exhibit varying alignment individually at different scales.  
It is necessary to constrain them during the training process to ensure alignment between image-text pair at various scale levels.

Considering a batch of image-text pairs as input, we can obtain image-text alignment similarity matrices at four different scales, after passing through the MSCMAT. We utilize the MSCMA loss to optimize these alignment matrices. This loss has the same form as the cross-modal contrastive loss at a single scale, as shown in Equation~\ref{eq14}. 

\begin{equation}
\begin{split}
L_{CMA}(\mathbf{m}^{i})
= -\frac{1}{2}[
 \frac{1}{b}\sum_{j=1}^{b}\log{\frac{exp\left({\mathbf{m}^{i}}_{jj}/\tau\right)}{\sum_{u=1}^{b}exp\left({\mathbf{m}^{i}}_{ju}/\tau\right)}} +\\
\frac{1}{b}\sum_{j=1}^{b}\log{\frac{exp\left({\mathbf{m}^{i}}^{\textbf{T}}_{jj}/\tau\right)}{\sum_{u=1}^{b}exp\left({\mathbf{m}^{i}}^{\textbf{T}}_{ju}/\tau\right)}}]
\end{split}
\label{eq14}
\end{equation}
The subscript of $\mathbf{m}^{i}$ represents the index of the matrix element and \textbf{T} represents the transpose of a matrix. The multi-scale cross-modal alignment loss encourages the model to maximize the similarity between positive pairs (matching image-text pairs) while minimizing the similarity between negative pairs (non-matching image-text pairs) at each scale. 

Then, we sum the $L_{CMA}$ at all scales together to obtain the overall MSCMA loss $L_{MSCMA}$, as shown in Equation~\ref{eq15}. 

\begin{equation}
L_{MSCMA}\ = \sum_{i=1}^{n}{L_{CMA}(\mathbf{m}^{i})}
\label{eq15}
\end{equation}

By optimizing this loss, the model effectively learns to align representations between image and text at different scales, achieving multi-scale cross-modal semantic alignment and enhancing the overall performance of the retrieval system.

\begin{figure*}[t]
\centering
\includegraphics[width=1\textwidth]{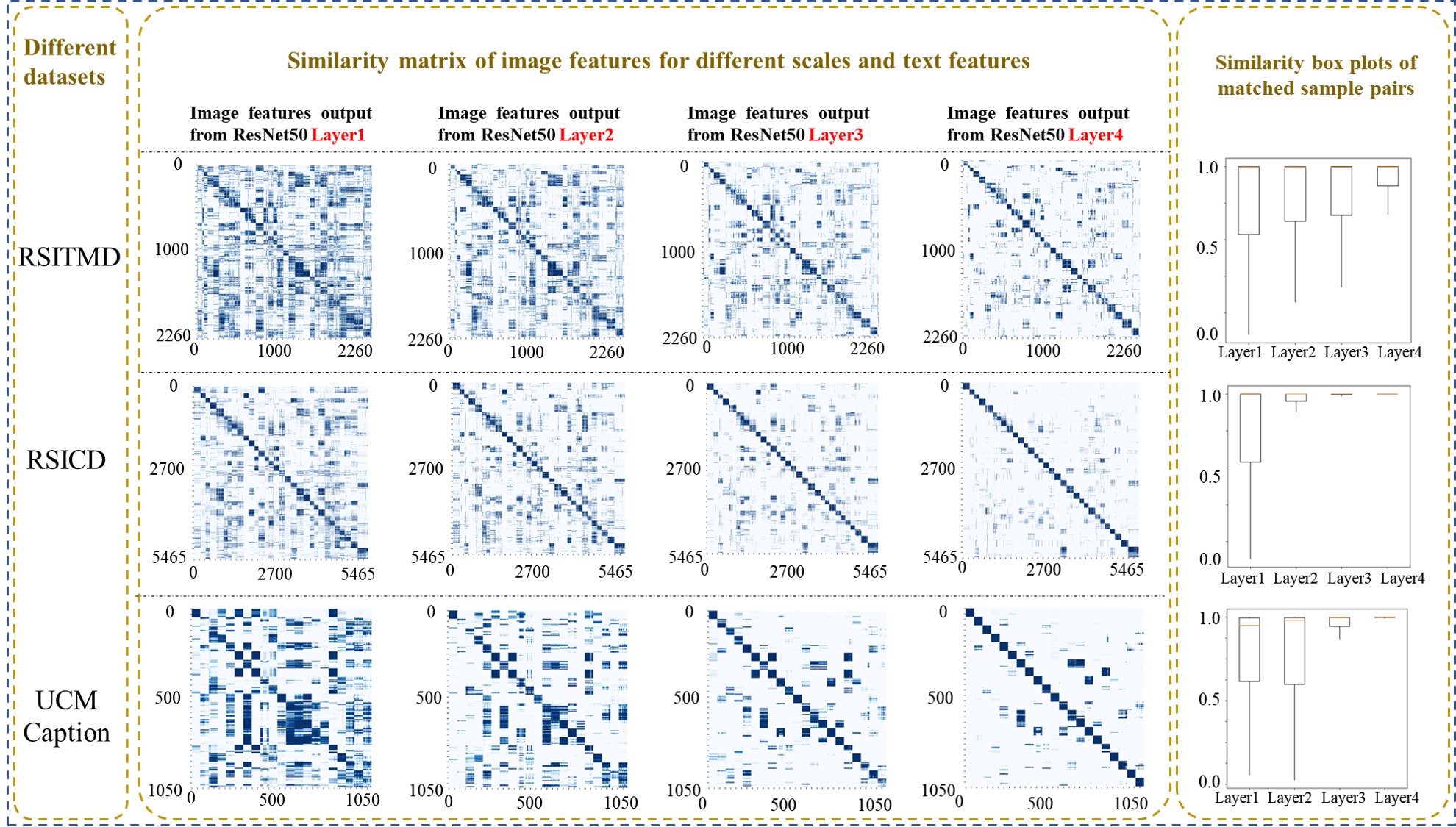} 
 \caption{There is a schema about lacking of cross-scale multi-modal semantic consistency. Each heatmap in Figure represents the image-text similarity matrix at a specific scale, based on the test set of a particular dataset. The horizontal axis of the heatmap represents text IDs in the test set, while the vertical axis represents image IDs. Additionally, boxplot statistics were performed on the diagonal elements obtained from the four heatmaps in each dataset. The boxplots in Figure depict different scales on the horizontal axis and the similarity value on the vertical axis. The observations from the heatmaps and boxplots indicate that image-text alignment is stronger for larger scales (Layer4), where positive sample pairs are closer together and negative sample pairs are further apart. However, alignment becomes weaker as the scale decreases (from Layer 4 to Layer 1).}
\label{csmmc}
\end{figure*}

\subsection{Cross-Scale Multi-Modal semantic Consistency loss (CSMMC loss)}

\label{Analysis of Cross-scale Multi-modal Consistency.}

During the separate image-text alignments at multi-scale, it was observed that the alignment difficulty varied across different scales. Statistical analysis was conducted on three RSITR datasets, where features from different layers of ResNet50 were aligned with text features and visually examined, as shown in Figure~\ref{csmmc}. We obtained image-text similarity matrices heatmap at different scales for all image-text pairs in three test sets. Furthermore, we performed boxplot statistics on the diagonal elements obtained from the four heatmaps in each test set. By observing Figure~\ref{csmmc}, we can find that the best alignment occurred in Layer 4 along the main diagonal, while the alignment becomes poorer for smaller scales (from Layer 4 to Layer 1).

This phenomenon indicates that larger-scale image features align more easily with text, while smaller-scale features present greater challenges due to their focus on lower-level image details such as textures and edges \cite{zeiler2014visualizing}. In this case, the alignment information from larger-scale features can be utilized to guide the alignment of smaller-scale features, enhancing cross-scale multi-modal semantic consistency. Therefore, a novel CSMMC loss was designed to achieve this, ensuring both the consistency of multi-modal semantics across scales and the diversity of original multi-scale image features. This loss leads to improved retrieval performance by learning more comprehensive joint representations of images and text.

Specifically, we take the image-text similarity matrix at the largest scale as the teacher information to minimize the K-L divergence between the distribution of the similarity matrices of the remaining scales and the largest scale, aiming to bridge the gap in alignment difficulty among different scales. The specific process is illustrated by the following equations:

\begin{equation}
\begin{split}
L_{MMC}(\mathbf{m}^{i},\mathbf{m}^{n})=\frac{1}{b}\sum_{j=1}^{b} \mathbf{KL}(Softmax(\frac{\mathbf{m}^{i}_{j}}{\mu}), \\Softmax(\frac{\mathbf{m}^{n}_{j}}{\mu}))
\end{split}
\label{eq16}
\end{equation}

\begin{equation}
L_{CSMMC}\ = \sum_{i=1}^{n}{L_{MMC}(\mathbf{m}^{i},\mathbf{m}^{n})}
\label{eq17}
\end{equation}
Here, $j$ represents the row index of $\mathbf{m}^i$ and $\mathbf{m}^n$. $\mathbf{m}^i$ refers to the image-text similarity matrix at the i-th scale, while $\mathbf{m}^n$ refers to the image-text similarity matrix at the largest scale.

It is worth noting that, while enhancing cross-scale multi-modal semantic consistency, we can ensure both the consistency of multi-modal semantics among different scales and the diversity of the original image features at different scales. This is because the constraint is applied to the similarity matrix, without directly constraining the original image features.

\subsection{Training and Testing}

The proposed MSA in this paper is an end-to-end training method. During training, the image encoder, text encoder, separate alignment module, and fusion module are jointly optimized under the constraint of the loss function as shown in Equation~\ref{eq8}. Since the MSCMAT achieves interaction between multi-scale image features and text features, the model is a single-stream model during training. During testing, the data flow does not pass through the MSCMAT. The image data is processed by the image encoder and the fusion module to obtain multi-scale fused image features. The text data is processed by BERT, and the $cls$ feature is used as the text feature. Only the cosine similarity between the multi-scale fused image features and text features needs to be calculated. Therefore, in testing, the proposed method in this paper is a dual-stream model that allows offline feature extraction for the data to be retrieved, making it deployable in practical retrieval scenarios.

\begin{algorithm}[h]
\caption{The training algorithm of proposed MSA}\label{alg1}
\begin{algorithmic}
\STATE \textbf{Input:}
\STATE \hspace{0.5cm} Training set:$D = \{(I_i,T_i)\}$, $i\in(1,n)$.
\STATE \hspace{0.5cm} $\alpha$, $\beta$, Learning rate lr, iterative epoch K.
\STATE \textbf{Output:}
\STATE \hspace{0.5cm} The parameters $\Theta_{img}$ of the image encoder, the parameters $\Theta_{text}$ of text encoder, the parameters $\Theta_{mscmat}$ of MSCMAT, the parameters $\Theta_{fuse}$ of the fusion module.

\STATE \textbf{Initialization:}
\STATE \hspace{0.5cm} The parameters $\Theta_{img}$ are loaded with pre-trained weights from the Resnet. $\Theta_{text}$ are loaded with pre-trained weights from the BERT-base.  $\Theta_{mscmat}$ and $\Theta_{fuse}$ randomly initialized by normal distribution. 
\STATE \textbf{Repeat:}
\STATE \hspace{0.5cm} Step 1: Sample the input RS image-text pairs $\left( {I_{i},T_{i}} \right)$;
\STATE \hspace{0.5cm} Step 2: Calculate the multi-scale image feature $\mathbf{v}^i$ via Equation~\ref{eq1} -~\ref{eq2} and text features via Equation~\ref{eq3};
\STATE \hspace{0.5cm} Step 3: Calculate the similarity matrix $\mathbf{m}^{i}$ for each scale via Equation~\ref{eq4};
\STATE \hspace{0.5cm} Step 4: Calculate the multi-scale fusion image features $\mathbf{v}^{\prime\prime}$ via Equation~\ref{eq5} and ~\ref{eq6};
\STATE \hspace{0.5cm} Step 5: Calculate the loss $L_{MSCMA}$ according to Equation~\ref{eq14} -~\ref{eq15};
\STATE \hspace{0.5cm} Step 6: Calculate the loss $L_{CSMMC}$ according to Equation~\ref{eq16} -~\ref{eq17};
\STATE \hspace{0.5cm} Step 7: Calculate the loss $L_{tri}$ according to Equation~\ref{eq7};
\STATE \hspace{0.5cm} Step 8: Calculate the totel loss $L_{totel}$ according to Equation~\ref{eq8};
\STATE \hspace{0.5cm} Step 9: Update $\Theta_{img}$, $\Theta_{text}$, $\Theta_{mscmat}$ and $\Theta_{fuse}$;
\STATE \textbf{Until:} Training reaches the K-th
epoch or the network has converged.
\STATE \textbf{Return:} The parameters $\Theta_{img}$, $\Theta_{text}$, $\Theta_{mscmat}$ and $\Theta_{fuse}$
\end{algorithmic}
\label{alg1}
\end{algorithm}

\section{Experiments}

\subsection{Datasets, Implementation details and Metrics}

We conducted experiments on three commonly used RSITR datasets: RSITMD \cite{yuan2021exploring}, RSICD \cite{lu2017exploring}, and UCM Caption \cite{qu2016deep}. The RSITMD dataset consists of 4,743 RS images distributed across 32 different scenes. The RSICD dataset contains 10,921 RS images encompassing 31 scenes. The UCM Caption dataset comprises 2,100 RS images across 21 scenes. Each image in these datasets is accompanied by five corresponding text sentences describing the content.

During the training process, we fixed the batch size to 32 and the learning rate to 0.00001. We utilized the Adam optimizer with optimizer parameters set to $weight\_decay=0.1,betas=(0.9, 0.999)$. Programming was carried out using PyTorch 1.7.1, and experiments were conducted on an Nvidia RTX 3090. 

Following previous work \cite{yuan2021exploring}, we evaluated the performance using the R@K metric ($K=1, 5, 10$) in our experiments. R@K represents the percentage of correctly retrieved results within the top $K$ positions, out of all retrieval attempts. We conducted image-to-text and text-to-image retrieval modes separately and obtained the three R@K values for each mode. The average of all the six R@K values from both retrieval modes is denoted as mR.

\subsection{Comparison with state-of-the-arts}

We compared the proposed method with several classical image-text retrieval methods (VSE++ \cite{faghri2017vse++}, SCAN \cite{lee2018stacked}, MTFN \cite{wang2019matching}, CAMP \cite{Wang_2019_ICCV}, CAMERA \cite{qu2020context}) as well as several RSITR methods (AMFMN \cite{yuan2021exploring}, GaLR \cite{yuan2022remote}, MCRN \cite{yuan2022mcrn}, CABIR \cite{zheng2022cross}, LW-MCR \cite{yuan2021lightweight}, HVSA \cite{zhang2023hypersphere}, HyperMatch \cite{yao2022hypergraph}, SWAN \cite{pan2023reducing}, SSJDN \cite{zheng2023scale}, MGRM \cite{zhang2023exploring}, SMLGN \cite{chen2024integrating}, KAMCL \cite{ji2023knowledge}, MSITA \cite{chen2023multiscale}).
Additionally, to ensure a fair comparison and eliminate the influence of the text encoder BERT, we also included several methods: ResNet + BERT \cite{he2016deep,devlin2018bert} and VIT + BERT \cite{dosovitskiy2020image}.

VSE++ is a method that utilizes the VGG network and GRU network as feature extractors and optimizes them using triplet loss. SCAN is a fine-grained alignment method based on stacked attention for image-text retrieval. MTFN is a matching method that maps images and text into multiple tensor spaces. CAMP is a fine-grained image-text matching approach. CAMERA is a context-aware and multi-view-based image-text matching method. 

AMFMN is an RSITR method that considers the multi-scale nature of remote sensing images. GaLR is a fusion method that combines local and global information. LW-MCR is a lightweight RSITR method. MCRN is a general framework that enables retrieval of RS images, text, and audio. HVSA is a hyper-spherical learning-based RS image-text matching network with curriculum learning. HyperMatch is a text-visual matching network based on hypergraph enhancement. SWAN is a method that combines multi-scale image feature fusion and text coarse-grained enhancement. SSJDN addresses the limitations of methods focused solely on scale or semantics. MGRM tackles information redundancy and accuracy degradation in current methods. SMLGN enhances the use of rich textual descriptions in RS images. KAMCL, a knowledge-aided approach, resolves the issue of overlooking subtle differences in highly similar RSITR descriptions. MSITA overcomes existing methods' failures in capturing crucial information and establishing prior similarity between images and texts.

ResNet + BERT and VIT + BERT utilize ResNet (18, 50, 101) or VIT as the image feature extractor and BERT as the text feature extractor, optimizing them using triplet loss.

Our method MSA was implemented with three variants, using ResNet101, ResNet50, and ResNet18 as the image encoders, denoted as MSA-101, MSA-50, and MSA-18, respectively. Table~\ref{tab:1}-\ref{tab:3} presents the experimental results obtained on the RSITMD, RSICD, and UCM Caption datasets, respectively.

\begin{table}[]

\caption{\label{tab:1}Experimental results on RSITMD}
\resizebox{\linewidth}{!}{
\begin{tabular}{lccccccc}
\hline
\multicolumn{1}{c}{\multirow{2}{*}{\textit{Method}}} & \multicolumn{3}{c}{\textit{Sentence   Retrieval}} & \multicolumn{3}{c}{\textit{Image   Retrieval}} & \multirow{2}{*}{\textit{mR}} \\
\multicolumn{1}{c}{}                                 & \textit{R@1}   & \textit{R@5}   & \textit{R@10}   & \textit{R@1}  & \textit{R@5}  & \textit{R@10}  &                              \\ \hline
\multicolumn{8}{c}{\textit{\textbf{Traditional methods}}}\\ \hline
VSE++ \cite{faghri2017vse++} \textit{BMVC'18}                                             & 9.07           & 21.61          & 31.78           & 7.73          & 27.80          & 41.00             & 23.17                        \\
SCAN-t2i \cite{lee2018stacked} \textit{ECCV'18}                                            & 10.18          & 28.53          & 38.49           & 10.10          & 28.98         & 43.53          & 26.64                        \\
SCAN-i2t \cite{lee2018stacked} \textit{ECCV'18}                                            & 11.06          & 25.88          & 39.38           & 9.82          & 29.38         & 42.12          & 26.28                        \\
CAMP-triplet \cite{Wang_2019_ICCV} \textit{ICCV'19}                                         & 11.73          & 26.99          & 38.05           & 8.27          & 27.79         & 44.34          & 26.20                         \\
CAMP-bce \cite{Wang_2019_ICCV} \textit{ICCV'19}                                             & 9.07           & 23.01          & 33.19           & 5.22          & 23.32         & 38.36          & 22.03                        \\
MTFN \cite{wang2019matching} \textit{MM'19}                                                & 10.4           & 27.65          & 36.28           & 9.96          & 31.37         & 45.84          & 26.92                        \\
CAMERA \cite{qu2020context} \textit{MM'20}                                              & 8.33           & 21.83          & 33.11           & 7.52          & 26.19         & 40.72          & 22.95                        \\ \hline
\multicolumn{8}{c}{\textit{\textbf{RSITR methods}}}    \\ \hline
AMFMN-soft \cite{yuan2021exploring} \textit{TGRS'22}                                          & 11.06          & 25.88          & 39.82           & 9.82          & 33.94         & 51.90           & 28.74                        \\
AMFMN-fusion \cite{yuan2021exploring} \textit{TGRS'22}                                         & 11.06          & 29.20           & 38.72           & 9.96          & 34.03         & 52.96          & 29.32                        \\
AMFMN-sim \cite{yuan2021exploring} \textit{TGRS'22}                                           & 10.63          & 24.78          & 41.81           & 11.51         & 34.69         & 54.87          & 29.72                        \\
MRCN \cite{yuan2022mcrn} \textit{JAG'22}                                                & 13.27          & 29.42          & 41.59           & 9.42          & 35.53         & 52.74          & 30.33                        \\
LW-MCR-b \cite{yuan2021lightweight} \textit{TGRS'22}                                            & 9.07           & 22.79          & 38.05           & 6.11          & 27.74         & 49.56          & 25.55                        \\
LW-MCR-d \cite{yuan2021lightweight} \textit{TGRS'22}                                           & 10.18          & 28.98          & 39.82           & 7.79          & 30.18         & 49.78          & 27.79                        \\
LW-MCR-u \cite{yuan2021lightweight} \textit{TGRS'22}                                        & 9.73           & 26.77          & 37.61           & 9.25          & 34.07         & 54.03          & 28.58                        \\
GALR \cite{yuan2021lightweight} \textit{TGRS'22}                                                & 13.05          & 30.09          & 42.70            & 10.47         & 36.34         & 53.35          & 31.00                    \\
GALR-MR \cite{yuan2021lightweight} \textit{TGRS'22}                                             & 14.82          & 31.64          & 42.48           & 11.15         & 36.68         & 51.68          & 31.41                        \\
SWAN \cite{pan2023reducing} \textit{ICMR'23}                                                & 13.35          & 32.15          & 46.90            & 11.24         & 40.40          & 60.60          & 34.11                        \\
HyperMatch \cite{yao2022hypergraph} \textit{JSTARS'22}                                        & 11.73          & 28.10           & 38.05           & 9.16          & 32.31         & 46.64          & 27.67                        \\
HVSA \cite{zhang2023hypersphere} \textit{TGRS'23}                                                 & 13.20           & 32.08          & 45.58           & 11.43         & 39.20          & 57.45          & 33.16                        \\ 
SSJDN \cite{zheng2023scale} \textit{ACM TOMM'23}                                                & 11.28          & 28.09        & 41.59          & 13.23         & 43.45          & \textbf{64.73}          &33.73                     \\ 
MGRM \cite{zhang2023exploring} \textit{TGRS'23}                                               & 13.51         &31.87       & 46.27         & 11.11         & 37.22         & 56.61         &32.76                    \\ 

SMLGN \cite{chen2024integrating} \textit{TGRS'24}                                              & 17.26         &39.38     & 51.55         &13.19       &43.94       & 60.40        &37.62                  \\ 

KAMCL \cite{ji2023knowledge} \textit{TGRS'23}                                               & 16.51        &36.28       & 49.12        & 13.50        & 42.15       & 59.32         &36.14                    \\ 

MSITA \cite{chen2023multiscale} \textit{TGRS'24}                                               & 15.22         &34.20      &47.65         & 12.15       & 39.92        &57.72      &34.48                \\ 

\hline
\multicolumn{8}{c}{\textit{\textbf{Additional variants}}}     \\ \hline
ResNet18 + BERT                                    & 13.94          & 32.08         & 46.02          &11.28        & 38.05        &56.64& 33.00                     \\
ResNet50 + BERT                                    & 17.70       & 36.06         & 46.02         &12.26         & 38.72        &54.03         & 34.13                       \\
ResNet101 + BERT                                    & 15.71           & 32.74           & 44.25           & 13.14         & 41.77        & 58.01         &34.27                      \\ 
VIT + BERT                                             & 14.16          & 34.07          & 46.46          & 13.19          & 43.10         & 60.13          & 35.18                        \\\hline
\multicolumn{8}{c}{\textit{\textbf{Ours}}} \\ \hline
MSA-101                                           & 15.93        & 38.50        &50.88          &14.96        & \textbf{45.22}         & 61.86         & 37.89            \\
MSA-50                                            & 16.59          & \textbf{40.04}        & \textbf{51.33}          &\textbf{15.53}        & 44.20        & 60.80          & \textbf{38.08}                       \\
MSA-18                                            & \textbf{19.25}        & 37.61         & 50.44           & 13.63         &40.80        & 57.92        &36.61  \\ \hline
\end{tabular}}
\end{table}

From Table~\ref{tab:1}, it can be observed that on the RSITMD dataset, our method MSA-50 achieved an mR value of 38.08, surpassing all the comparison methods. MSA-101 and MSA-18 also achieved the second and third-highest mR values, respectively. Compared to some methods that consider multiple-scale nature, such as AMFMN, GaLR, and SWAN, our method achieved the highest mR even with the use of ResNet18 as the visual backbone. While GaLR and SWAN utilized FastRcnn and ResNet50, along with complex filtering and fusion mechanisms. Some methods, like HVSA and SSJDN, sacrificed some R@1 to improve overall performance, whereas our method achieved high values for both R@1 and mR.
Furthermore, to eliminate the influence of the text feature extraction network, we compared our proposed method with ResNet18 + BERT, ResNet50 + BERT, ResNet101 + BERT, and found that MSA-101 outperformed ResNet101+BERT by 3.62, MSA-50 outperformed ResNet50 + BERT by 3.95, and MSA-18 outperformed ResNet18 + BERT by 3.61. Even MSA-18 outperformed ResNet101 + BERT and VIT + BERT by 2.34 and 1.43, respectively. This indicates that our MSA method while enhancing the representation through multi-scale cross-modal semantic alignment and cross-scale multi-modal semantic consistency is highly effective in improving retrieval performance for RS images.

\begin{table}[]

\caption{\label{tab:2}Experimental results on RSICD}
\resizebox{\linewidth}{!}{
\begin{tabular}{lccccccc}
\hline
\multirow{2}{*}{\textit{Method}} & \multicolumn{3}{c}{\textit{Sentence   Retrieval}} & \multicolumn{3}{c}{\textit{Image   Retrieval}} & \multirow{2}{*}{\textit{mR}} \\
                                 & \textit{R@1}   & \textit{R@5}   & \textit{R@10}   & \textit{R@1}  & \textit{R@5}  & \textit{R@10}  &                              \\ \hline
\multicolumn{8}{c}{\textit{\textbf{Traditional methods}}}  \\ \hline
VSE++ \cite{faghri2017vse++} \textit{BMVC'18}                           & 4.56           & 16.73          & 22.94           & 4.37          & 15.37         & 25.35          & 14.89                        \\
SCAN-t2i \cite{lee2018stacked} \textit{ECCV'18}                        & 4.39           & 10.90           & 17.64           & 3.91          & 16.20          & 26.49          & 13.25                        \\
SCAN-i2t \cite{lee2018stacked} \textit{ECCV'18}                         & 5.85           & 12.89          & 19.84           & 3.71          & 16.40          & 26.73          & 14.23                        \\
CAMP-triplet \cite{Wang_2019_ICCV} \textit{ICCV'19}                     & 5.12           & 12.89          & 21.12           & 4.15          & 15.23         & 27.81          & 14.39                        \\
CAMP-bce \cite{Wang_2019_ICCV} \textit{ICCV'19}                         & 4.20            & 10.24          & 15.45           & 2.72          & 12.76         & 22.89          & 11.38                        \\
MTFN \cite{wang2019matching} \textit{MM'19}                            & 5.02           & 12.52          & 19.74           & 4.90           & 17.17         & 29.49          & 14.81                        \\
CAMERA \cite{qu2020context} \textit{MM'20}                            & 4.57           & 13.08          & 21.77           & 4.00             & 15.93         & 26.97          & 14.39                        \\ \hline
\multicolumn{8}{c}{\textit{\textbf{RSITR methods}}}        \\ \hline
AMFMN-soft \cite{yuan2021exploring} \textit{TGRS'22}                      & 5.05           & 14.53          & 21.57           & 5.05          & 19.74         & 31.04          & 16.02                        \\
AMFMN-fusion \cite{yuan2021exploring} \textit{TGRS'22}                     & 5.39           & 15.08          & 23.40            & 4.90           & 18.28         & 31.44          & 16.42                        \\
AMFMN-sim \cite{yuan2021exploring} \textit{TGRS'22}                       & 5.21           & 14.72          & 21.57           & 4.08          & 17.00            & 30.60           & 15.53                        \\
MRCN \cite{yuan2022mcrn} \textit{JAG'22}                            & 6.59           & 19.40           & 30.28           & 5.03          & 19.38         & 32.99          & 18.95                        \\
LW-MCR-b \cite{yuan2021lightweight} \textit{TGRS'22}                        & 4.57           & 13.71          & 20.11           & 4.02          & 16.47         & 28.23          & 14.52                        \\
LW-MCR-d \cite{yuan2021lightweight} \textit{TGRS'22}                         & 3.29           & 12.52          & 19.93           & 4.66          & 17.51         & 30.02          & 14.66                        \\
LW-MCR-u \cite{yuan2021lightweight} \textit{TGRS'22}                        & 4.39           & 13.35          & 20.29           & 4.30           & 18.85         & 32.34          & 15.59                        \\
GALR \cite{yuan2021lightweight} \textit{TGRS'22}                            & 6.50            & 18.91          & 29.70            & 5.11          & 19.57         & 31.92          & 18.62                        \\
GALR-MR \cite{yuan2021lightweight} \textit{TGRS'22}                         & 6.59           & 19.85          & 31.04           & 4.69          & 19.48         & 32.13          & 18.96                        \\
CABIR \cite{zheng2022cross} \textit{AS'22}                            & 8.59           & 16.27          & 24.13           & 5.42          & 20.77         & 33.58          & 18.12                        \\
SWAN \cite{pan2023reducing} \textit{ICMR'23}                            & 7.41           & 20.13          & 30.86           & 5.56          & \textbf{22.26}         & 37.41          & 20.61                        \\
HyperMatch \cite{yao2022hypergraph} \textit{JSTARS'22}                       & 7.14           & 20.04          & 31.02           & 6.08          & 20.37         & 33.82          & 19.75                        \\
HVSA \cite{zhang2023hypersphere} \textit{TGRS'23}                            & 7.47           & 20.62          & 32.11           & 5.51          & 21.13         & 34.13          & 20.16                        \\
SSJDN \cite{zheng2023scale} \textit{ACM TOMM'23}                                                 &  7.69         &20.59        &  32.20         & 5.58        & 22.07         &36.54           &    20.78               \\ 
MGRM \cite{zhang2023exploring} \textit{TGRS'23}                                               &   7.41     & 23.24    & 35.32       & 5.75        &  21.23      &35.55        &  21.42              \\ 

MSITA \cite{chen2023multiscale} \textit{TGRS'24} &                                        8.67 &  22.71   &  33.91    &   6.13     &   21.98    & 35.39       & 21.47             \\

\hline
\multicolumn{8}{c}{\textit{\textbf{Additional variants}}}   \\ \hline
ResNet18 + BERT & 7.87 & 22.69 & 33.94 & 5.93 & 22.69 & 38.5  & 21.94 \\
ResNet50 + BERT & 7.96 & 23.33 & 32.57 & 6.08 & 21.68 & 37.16 & 21.46 \\
ResNet101 + ERT & 8.51 & 23.15 & 35.77 & 5.62 & 21.87 & 38.13 & 22.17 \\
\hline
\multicolumn{8}{c}{\textit{\textbf{Ours}}}                   \\ \hline
MSA-101 &  \textbf{9.52}    & \textbf{25.25}      &      35.32 &   6.55   &   \textbf{24.43}    &  \textbf{39.63}     &  \textbf{23.45}     \\
MSA-50  & 7.50  & 23.70  & \textbf{36.96} & \textbf{6.77} & 23.04 & 39.03 & 22.83 \\
MSA-18  & 7.96 & 23.33 & 35.77 & 5.97 & 22.93 & 39.58 & 22.59                       \\ \hline
\end{tabular}}
\end{table}

From Table~\ref{tab:2}, it can be observed that on the RSICD dataset, MSA-101, MSA-50, and MSA-18 achieved mR values of 23.45, 22.83, and 22.59, respectively, surpassing all the comparison algorithms. 
We compared our proposed method with ResNet18 + BERT, ResNet50 + BERT, ResNet101 + BERT, and found that under the same feature extraction networks, our method demonstrated better performance, achieving mR improvements of 1.28, 1.37, and 0.65, respectively. Even when comparing MSA-18 with ResNet101 + BERT, our method achieved an mR improvement of 0.42. This indicates that our method of aligning multiple scales of image features and text features separately is effective.

\begin{table}[]

\caption{\label{tab:3}Experimental results on UCM Caption}
\resizebox{\linewidth}{!}{
\begin{tabular}{lccccccc}
\hline
\multicolumn{1}{c}{\multirow{2}{*}{\textit{Method}}} & \multicolumn{3}{c}{Sentence   Retrieval}                                                                & \multicolumn{3}{c}{Image   Retrieval}                                                                   & \multicolumn{1}{c}{\multirow{2}{*}{\textit{mR}}} \\
\multicolumn{1}{c}{}                                 & \multicolumn{1}{c}{\textit{R@1}} & \multicolumn{1}{c}{\textit{R@5}} & \multicolumn{1}{c}{\textit{R@10}} & \multicolumn{1}{c}{\textit{R@1}} & \multicolumn{1}{c}{\textit{R@5}} & \multicolumn{1}{c}{\textit{R@10}} & \multicolumn{1}{c}{}                             \\ \hline
\multicolumn{8}{c}{\textit{\textbf{Traditional methods}}}      \\ \hline
VSE++ \cite{faghri2017vse++} \textit{BMVC'18}                                                & 12.38                            & 44.76                            & 65.71                             & 10.10                            & 31.80                            & 56.85                             & 36.93                                            \\
SCAN-t2i \cite{lee2018stacked} \textit{ECCV'18}                                            & 14.29                            & 45.71                            & 67.62                             & 12.76                            & 50.38                            & 77.24                             & 44.67                                            \\
SCAN-i2t \cite{lee2018stacked} \textit{ECCV'18}                                              & 12.85                            & 47.14                            & 69.52                             & 12.48                            & 46.86                            & 71.71                             & 43.43                                            \\
CAMP-triplet \cite{Wang_2019_ICCV} \textit{ICCV'19}                                          & 10.95                            & 44.29                            & 65.71                             & 9.90                             & 46.19                            & 76.29                             & 42.22                                            \\
CAMP-bce \cite{Wang_2019_ICCV} \textit{ICCV'19}                                              & 14.76                            & 46.19                            & 67.62                             & 11.71                            & 47.24                            & 76.00                             & 43.92                                            \\
MTFN \cite{wang2019matching} \textit{MM'19}                                                & 10.47                            & 47.62                            & 64.29                             & 14.19                            & 52.38                            & 78.95                             & 44.65                                            \\ \hline
\multicolumn{8}{c}{\textit{\textbf{RSITR methods}}}      \\ \hline
AMFMN-soft \cite{yuan2021exploring} \textit{TGRS'22}                                           & 12.86                            & 51.90                            & 66.76                             & 14.19                            & 51.71                            & 78.48                             & 45.97                                            \\
AMFMN-fusion \cite{yuan2021exploring} \textit{TGRS'22}                                         & 16.67                            & 45.71                            & 68.57                             & 12.86                            & 53.24                            & 79.43                             & 46.08                                            \\
AMFMN-sim \cite{yuan2021exploring} \textit{TGRS'22}                                            & 14.76                            & 49.52                            & 68.10                             & 13.43                            & 51.81                            & 76.48                             & 45.68                                            \\
LW-MCR-b \cite{yuan2021lightweight} \textit{TGRS'22}                                            & 12.38                            & 43.81                            & 59.52                             & 12.00                            & 46.38                            & 72.48                             & 41.10                                            \\
LW-MCR-d \cite{yuan2021lightweight} \textit{TGRS'22}                                            & 15.24                            & 51.90                            & 62.86                             & 11.90                            & 50.95                            & 75.24                             & 44.68                                            \\
LW-MCR-u \cite{yuan2021lightweight} \textit{TGRS'22}                                             & \textbf{18.10}                            & 47.14                            & 63.81                             & 13.14                            & 50.38                            & 79.52                             & 45.35                                            \\
CABIR \cite{zheng2022cross} \textit{AS'22}                                               & 15.17                            & 45.71                            & 72.85                             & 12.67                            & 54.19                            & 89.23                             & 48.30                                            \\ 
 SSJDN \cite{zheng2023scale} \textit{ACM TOMM'23} &  17.86 & 53.57 & 72.02 &  20.54 &  \textbf{62.56} &  82.98 & 51.59 \\

 SMLGN \cite{chen2024integrating} \textit{TGRS'24}                                              &  12.86                         &  49.52                            & 75.71  & 14.29                         &  52.76                         &84.67                           &48.30   

\\
 MSITA \cite{chen2023multiscale} \textit{TGRS'24} &                                       16.86  &   49.33  &   73.33    &   14.29  &   57.16    &  91.58     &  50.43        \\

\hline
\multicolumn{8}{c}{\textit{\textbf{Additional variants}}}                                                                                                          \\ \hline
ResNet18 + BERT   & 16.67 & 44.29 & 72.38 & 12.29 & 52.00    & 79.81 & 46.24 \\
ResNet50 + BERT   & 13.81 & 52.86 & 76.19 & 12.19 & 53.05 & \textbf{92.48} & 50.10  \\
ResNet101+ BERT   & 14.29 & 50.48 & 74.29 & 13.90 & 58.00    & 86.95 & 49.65 \\
VIT+BERT          & 10.95 & 44.29 & 70.00 & \textbf{14.86} & 55.90 & 90.95 & 47.83\\\hline
\multicolumn{8}{c}{\textit{\textbf{Ours}}}          \\ \hline
MSA-101          & 13.33 & \textbf{59.05} & \textbf{77.14} & 14.19 & 57.33 & 87.05 & 51.35 \\
MSA-50           & 13.33 & \textbf{59.05} & \textbf{77.14} & 13.14 & 57.52 & \textbf{92.48} & \textbf{52.11} \\
MSA-18           & 15.24 & 48.57 & 70.48 & 11.52 & 53.24 & 89.62 & 48.11                                          \\ \hline
\end{tabular}}
\end{table}

On the UCM dataset, we compared our method with previously reported results on the same dataset. MSA-101 and MSA-50 achieved mR values of 51.35 and 52.11, respectively, surpassing all the comparison methods. MSA-18 achieved an mR value of 48.11, only ranked after some latest methods, such as CABIR, SSJDN, and MSITA. However, these methods utilize a more complex backbone, such as ResNet152 as the visual backbone and BERT and bidirectional GRU as the text backbones, making it significantly more complex than the backbone of MSA-18. When using a larger visual backbone, MSA-50 and MSA-101 also outperformed these latest methods in terms of mR values. Therefore, the results on this dataset also demonstrate the superiority of our method. Additionally, MSA-101, MSA-50, and MSA-18 also outperformed ResNet101 + BERT, ResNet50 + BERT, and ResNet18 + BERT, respectively, demonstrating the effectiveness of our proposed method in enhancing multi-scale cross-modal semantic alignment and cross-scale multi-modal semantic consistency.

\subsection{Ablation Studies}

We conducted ablation experiments on the RSITMD and UCM Caption datasets to assess the effectiveness of the key innovations proposed in this paper. The ablation experiment consists of four parts. Firstly, we performed ablation on the key components proposed in this paper to investigate the effectiveness of multi-scale alignment. Subsequently, we carried out ablation on the structures of different multi-scale cross-modal alignment Transformers. Then, we investigated the impact of using MSCMAT at different layers of the image encoder on retrieval results. Lastly, we perform a sensitivity analysis of the hyperparameters.
\subsubsection{Ablation of the proposed components in this paper}

To explore the effectiveness of multi-scale alignment, we conducted various comparisons with different settings.
First, we established a Baseline (referred to as Base for convenience). The Base used the same feature extractor and triplet loss as our proposed method, without considering multi-scale features. Specifically, we used the flattened vector before the classifier of ResNet as the image feature and the output vector of the "cls" position in BERT as the text feature. The triplet loss was applied in the Base.

Based on the Baseline, we also constructed several variants: Baseline + Multi-scale Fusion (Base+M), Baseline + Multi-scale Fusion + Multi-scale Cross-modal Transformer + MSCMA loss (Base+M+A+B), and Baseline + Multi-scale Fusion + Multi-scale Cross-modal Transformer + MSCMA loss + CSMMC loss (Base+M+A+B+C, which represents our proposed method MSA).

To verify if our method is effective with various visual backbones, we conducted experiments on ResNet18, ResNet50, and ResNet101, evaluating the four settings mentioned above. The results are shown in Table~\ref{tab:4} to~\ref{tab:6}.

\begin{table}[]

\caption{\label{tab:4}Ablation results based on ResNet18}
\resizebox{\linewidth}{!}{
\begin{tabular}{cccccccc}
\hline
\multirow{2}{*}{\textit{Method}} & \multicolumn{3}{c}{Sentence   Retrieval}    & \multicolumn{3}{c}{Image   Retrieval}       & \multirow{2}{*}{\textit{mR}} \\
                                 & \textit{R@1} & \textit{R@5} & \textit{R@10} & \textit{R@1} & \textit{R@5} & \textit{R@10} &                              \\ \hline
\multicolumn{8}{c}{\textit{UCM Caption}}     \\ \hline
Base         & \textbf{16.67} & 44.29 & 72.38 & \textbf{12.29} & 52.00 & 79.81 & 46.24 \\
Base+M       & 11.43 & 47.14 & 70.48 & 11.71 & 52.10 & 86.19 & 46.51 \\
Base+M+A+B   & 10.00 & \textbf{50.48} & \textbf{72.86} & 11.05 & 50.38 & 88.38 & 47.19 \\
Base+M+A+B+C (MSA) & 15.24 & 48.57 & 70.48 & 11.52 & \textbf{53.24 }& \textbf{89.62} & \textbf{48.11} \\ \hline
\multicolumn{8}{c}{\textit{RSITMD}}      \\ \hline

Base         & 13.94 & 32.08 & 46.02 & 11.28 & 38.05 & 56.64 & 33.00 \\
Base+M       & 11.95 & 29.86 & 42.26 & 9.73  & 37.43 & 55.84 & 31.18 \\
Base+M+A+B   & 17.48 & 36.28 & 49.34 & \textbf{13.67} & \textbf{41.42} & \textbf{60.49} & 36.45 \\
Base+M+A+B+C (MSA) & \textbf{19.25} & \textbf{37.61} & \textbf{50.44} & 13.63 & 40.80 & 57.92 & \textbf{36.61}           \\ \hline
\end{tabular}}
\end{table}

From Table~\ref{tab:4}, we can observe that on the UCM dataset, the Base achieved an mR value of 46.24. Comparatively, the Base+M increased by 0.27 compared to the Base, indicating that directly using multi-scale fusion slightly improves the retrieval performance. The Base+M+A+B, compared to the Base and Base+M, achieved mR value improvements of 0.68 and 0.95, respectively. This suggests that aligning image features and text features at different scales separately to enhance multi-scale cross-modal alignment of image-text representations is effective. The proposed MSCMAT and MSCMA loss in this paper contribute to improving the joint representation of image-text and, consequently, enhancing retrieval performance.
Furthermore, the Base+M+A+B+C, compared to the Base, Base+M, and Base+M+A+B, achieved mR value improvements of 1.87, 1.60, and 0.92, respectively. This demonstrates the effectiveness of the proposed CSMMC loss, which effectively bridges the difficulty gap in image-text alignment across different scales, providing more supervised information for learning image features at multiple scales and thereby enhancing the joint representation and improving retrieval performance.
On the RSITMD dataset, the performance of the different settings follows a similar trend as observed on the UCM Caption dataset. The Base+M did not show an improvement in mR compared to the Baseline, while the Base+M+A+B resulted in an improvement of 3.45 over the Base. The mR value of Base+M+A+B+C reached 36.61, which is an improvement of 0.16 compared to Base+M+A+B. This indicates that the three key components proposed in this paper are effective on the RSITMD dataset, and the multi-scale cross-modal semantic alignment and cross-scale multi-modal semantic consistency are crucial for enhancing the joint representation of image-text pairs.

\begin{table}[]

\caption{\label{tab:5}Ablation results based on ResNet50}
\resizebox{\linewidth}{!}{
\begin{tabular}{cccccccc}
\hline
\multirow{2}{*}{\textit{Method}} & \multicolumn{3}{c}{Sentence Retrieval}      & \multicolumn{3}{c}{Image Retrieval}         & \multirow{2}{*}{\textit{mR}} \\
                                 & \textit{R@1} & \textit{R@5} & \textit{R@10} & \textit{R@1} & \textit{R@5} & \textit{R@10} &                              \\ \hline
\multicolumn{8}{c}{\textit{UCM Caption}}                                                                                                                    \\ \hline
Base         & 13.81 & 52.86 & 76.19 & 12.19 & 53.05 & 92.48 & 50.10                     \\
Base+M       & \textbf{16.19} & 53.33 & 75.71 & \textbf{14.57} & 53.52 & 84.19 & 49.59                     \\
Base+M+A+B   & 13.33 & 50.00 & 76.19 & 12.76 & 57.14 & \textbf{92.95} & 50.40                     \\
Base+M+A+B+C (MSA) & 13.33 & \textbf{59.05} & \textbf{77.14} & 13.14 & \textbf{57.52} & 92.48 & \textbf{52.11}                     \\ \hline
\multicolumn{8}{c}{\textit{RSITMD}}      \\ \hline
Base         & 17.70 & 36.06 & 46.02 & 12.26 & 38.72 & 54.03 & 34.13 \\
Base+M       & 14.60 & 32.52 & 48.00 & 12.12 & 41.81 & 59.38 & 34.74 \\
Base+M+A+B   & \textbf{18.14} & 36.95 & 50.22 & 15.00 & \textbf{45.00} & \textbf{61.02} & 37.72 \\
Base+M+A+B+C (MSA) & 16.59 & \textbf{40.04} & \textbf{51.33} & \textbf{15.53} & 44.20 & 60.80 & \textbf{38.08}                       \\ \hline                        
\end{tabular}}
\end{table}

Table~\ref{tab:5} presents the ablation experiment results using ResNet50 as the visual backbone. On the UCM dataset, the Base+M did not achieve an mR improvement compared to the Base. This indicates that using the multi-scale fusion strategy alone does not yield better image representations for the RSITR task. The Base+M+A+B achieved an mR value of 50.40, which is an improvement of 0.30 over the Baseline. The highest mR value of 52.11 was obtained by the Base+M+A+B+C, indicating the necessity of enhancing cross-scale multi-modal semantic consistency.
On the RSITMD dataset, we can similarly observe that the Base+M  shows a slight mR improvement compared to the Base. The Base+M+A+B reached an mR value of 37.72, which is an improvement of 3.59 over the Base. This demonstrates the effectiveness of the proposed method for enhancing cross-scale multi-modal semantic consistency. The Base+M+A+B+C achieved an mR improvement of 0.36 compared to the Baseline+M+A+B, highlighting the effectiveness of enhancing cross-scale multi-modal semantic consistency.

\begin{table}[]

\caption{\label{tab:6}Ablation results based on ResNet101}
\resizebox{\linewidth}{!}{
\begin{tabular}{cccccccc}
\hline
\multirow{2}{*}{\textit{Method}} & \multicolumn{3}{c}{Sentence Retrieval}      & \multicolumn{3}{c}{Image Retrieval}         & \multirow{2}{*}{\textit{mR}} \\
                                 & \textit{R@1} & \textit{R@5} & \textit{R@10} & \textit{R@1} & \textit{R@5} & \textit{R@10} &                              \\ \hline
\multicolumn{8}{c}{UCM Caption}                                                                                                                             \\ \hline
Base         & 14.29 & 50.48 & 74.29 & 13.90 & 58.00 & 86.95 & 49.65 \\
Base+M       & 8.57  & 47.14 & 69.52 & 13.14 & 54.95 & \textbf{91.33} & 47.44 \\
Base+M+A+B   & \textbf{15.24} & 54.76 & \textbf{80.00} & \textbf{14.48} & 55.71 & 87.52 & 51.29 \\
Base+M+A+B+C (MSA) & 13.33 & \textbf{59.05} & 77.14 & 14.19 & \textbf{57.33} & 87.05 & \textbf{51.35} \\ \hline
\multicolumn{8}{c}{RSITMD}                                                                                                                                  \\ \hline
Base         & 15.71 & 32.74 & 44.25 & 13.14 & 41.77 & 58.01 & 34.27 \\
Base+M       & 13.72 & 35.84 & 48.01 & 14.34 & 44.61 & 60.13 & 36.11 \\
Base+M+A+B   & \textbf{17.92} & 37.83 & 49.78 & \textbf{16.95} & 44.07 & 59.29 & 37.64 \\
Base+M+A+B+C (MSA)& 15.93 & \textbf{38.50} & \textbf{50.88} & 14.96 & \textbf{45.22} & \textbf{61.86} & \textbf{37.89}                        \\ \hline
\end{tabular}}
\end{table}

Table~\ref{tab:6} presents the ablation results using ResNet101 as the visual backbone. It can be observed that on the UCM dataset, the Base+M+A+B achieved an mR improvement of 1.64 compared to the Base, while the Base+M+A+B+C achieved an mR improvement of 0.06 compared to the Base+M+A+B. On the RSITMD dataset, the Base+M+A+B showed an mR improvement of 3.37 compared to the Base. Additionally, the Base+M+A+B+C further achieved an mR improvement of 0.25 compared to the Base+M+A+B. These results demonstrate the effectiveness of enhancing both multi-scale cross-modal semantic alignment and cross-scale multi-modal semantic consistency as proposed in this paper.

Based on the above analysis, we can draw the following conclusions:

(1) The proposed alignment of image features and text separately at multiple scales is more effective than aligning only the fused image features and text.
(2) The key components proposed in this study enhance the multi-scale cross-modal semantic alignment and cross-scale multi-modal semantic consistency in the joint representation of multi-scale image-text pairs, thereby improving retrieval performance.
(3) The proposed method is effective on different visual backbones, demonstrating the transferability and generalizability of our method.

\subsubsection{Impact of Structure of MSCMAT}

\begin{figure}[t]
\centering
\includegraphics[width=0.5\textwidth]{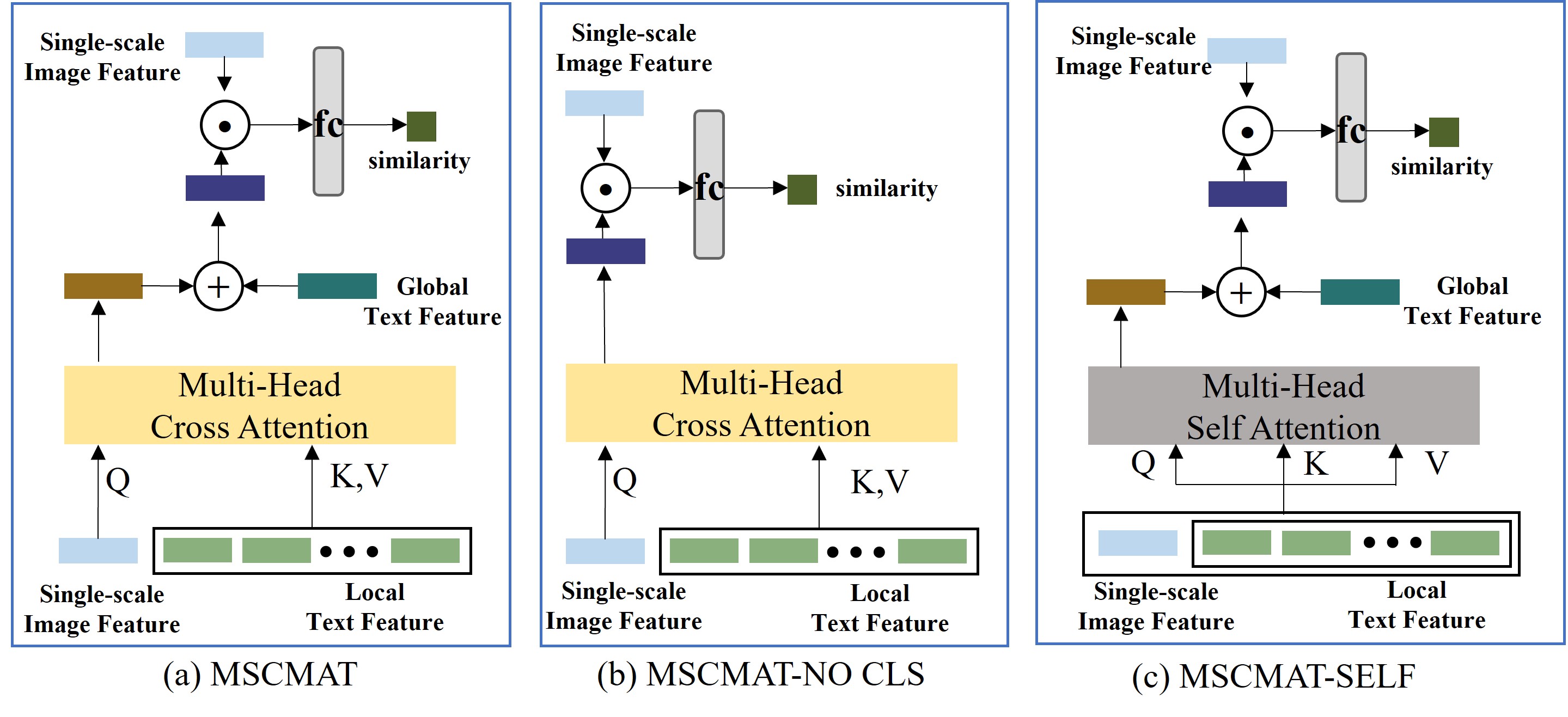} 
 \caption{Different structures of MSCMAT.}
\label{structure}
\end{figure}

We conducted ablation experiments on different structures of the MSCMAT. In addition to the proposed MSCMAT, we introduced two variants as shown in Figure~\ref{structure}. "MSCMAT-NO CLS" refers to using the Multi-Scale Cross-Modal Alignment Transformer without adding the CLS feature of the text, and MSCMAT-SELF refers to concatenating the image features with all local text features, followed by computing multi-head self-attention. We performed experiments on the RSITMD dataset using ResNet50 as the visual backbone, and the results are shown in Table~\ref{tab7}.

\begin{table}[]

\caption{\label{tab7}Results of different MSCMAT structures on ResNet-50}
\resizebox{\linewidth}{!}{
\begin{tabular}{cccccccc}
\hline
\multirow{2}{*}{\textit{Structure}} & \multicolumn{3}{c}{Sentence Retrieval}      & \multicolumn{3}{c}{Image Retrieval}         & \multirow{2}{*}{\textit{mR}} \\
                                    & \textit{R@1} & \textit{R@5} & \textit{R@10} & \textit{R@1} & \textit{R@5} & \textit{R@10} &                              \\ \hline
MSCMAT-SELF   & 17.04 & 36.5  & 48.23 & 14.87 & 42.3  & 59.07 & 36.33 \\
MSCMAT-NO-CLS & 17.48 & 38.94 & 50.66 & 15    & 44.56 & 61.64 & 38.05 \\
MSCMAT        & 16.59 & 40.04 & 51.33 & 15.53 & 44.2  & 60.8  & \textbf{38.08}       \\ \hline
\end{tabular}}
\end{table}

From Table~\ref{tab7}, we can observe that MSCMAT achieved an mR improvement of 1.75 compared to SELF. This indicates that using the cross-attention mechanism with image features as the Q vector is more effective in capturing the interactive information between images and text compared to using the self-attention mechanism. MSCMAT also achieved an mR improvement of 0.03 compared to MSCMAT-NO CLS. This suggests that introducing global text features after computing cross-attention to obtain aggregated text vectors can slightly aid in aligning single-scale images and text.

\subsubsection{Impact of Alignment at Different Layers}

To validate the effectiveness of separately aligning multi-scale image features with text information at different layers, we conducted alignment or non-alignment settings on four convolutional layers of ResNet50. We explored all possible combinations, resulting in a total of 15 different settings.

The experimental results are shown in Table~\ref{LAYERS}. From Table~\ref{LAYERS}, we can observe that Setting 1 (aligning image-text at all four scales) achieved the mR value of 38.08. Setting 2, Setting 3, Setting 5, and Setting 9, which aligned image-text at three scales, reached mR values of 36.84, 35.55, 37.23, and 36.32, respectively. On the other hand, Setting 4, Setting 6, Setting 7, Setting 10, Setting 11, and Setting 13, which aligned image-text at two scales, achieved mR values for 36.87, 36.41, 35.03, 35.59, 35.90, and 36.80, respectively. Lastly, Setting 8, Setting 12, Setting 14, and Setting 15, which aligned image-text at only one scale, obtained mR values of 34.92, 35.94, 35.66, and 35.28, respectively.

By observing the results, we can draw the following conclusions:
(1) Overall, the more layers of alignment, the better the performance. This confirms that alignment at different scales is beneficial, consistent with our previous analysis.
(2) When aligning image-text at all scales, the model achieved the optimal mR value. This demonstrates the effectiveness of separately aligning at all scales, validating the rationale behind our proposed method.

\begin{table}[t]
\caption{\label{tab:8}Ablation results on different layers for alignment}
\resizebox{\linewidth}{!}{
\begin{tabular}{lcccclllllll}
\hline
\multirow{2}{*}{Setting ID} & \multicolumn{4}{c}{\textit{Alignment in Layers}}                                          & \multicolumn{3}{c}{Sentence Retrieval} & \multicolumn{3}{c}{\textit{Image Retrieval}} & \multicolumn{1}{c}{\multirow{2}{*}{mR}} \\
                            & layer1               & layer2               & layer3               & layer4               & R@1         & R@5         & R@10       & R@1           & R@5           & R@10         & \multicolumn{1}{c}{}                    \\ \hline
1                                               & \checkmark & \checkmark & \checkmark & \checkmark & 16.59       & 40.04       & 51.33      & 15.53      & 44.20      & 60.80     & \textbf{38.08}               \\
2                                               & \checkmark & \checkmark & \checkmark & \multicolumn{1}{l}{}      & 16.81       & 36.73       & 48.89      & 12.65      & 43.63      & 62.35     & 36.84               \\
3                                               & \checkmark & \checkmark & \multicolumn{1}{l}{}      & \checkmark & 15.93       & 35.62       & 45.35      & 13.01      & 42.57      & 60.84     & 35.55               \\
4                                               & \checkmark & \checkmark & \multicolumn{1}{l}{}      & \multicolumn{1}{l}{}      & 17.04       & 37.39       & 49.56      & 13.81      & 43.81      & 59.60     & 36.87               \\
5                                               & \checkmark & \multicolumn{1}{l}{}      & \checkmark & \checkmark & 17.48       & 36.73       & 50.00      & 15.13      & 42.08      & 61.99     & 37.23               \\
6                                               & \checkmark & \multicolumn{1}{l}{}      & \checkmark & \multicolumn{1}{l}{}      & 18.58       & 35.62       & 47.12      & 14.03      & 42.96      & 60.13     & 36.41               \\
7                                               & \checkmark & \multicolumn{1}{l}{}      & \multicolumn{1}{l}{}      & \checkmark & 14.16       & 33.19       & 48.45      & 13.67      & 41.77      & 58.94     & 35.03               \\
8                                               & \checkmark & \multicolumn{1}{l}{}      & \multicolumn{1}{l}{}      & \multicolumn{1}{l}{}      & 15.93       & 34.51       & 46.46      & 13.72      & 41.73      & 57.17     & 34.92               \\
9                                               & \multicolumn{1}{l}{}      & \checkmark & \checkmark & \checkmark & 15.93       & 37.83       & 48.45      & 13.76      & 42.61      & 59.34     & 36.32               \\
10                                              & \multicolumn{1}{l}{}      & \checkmark & \checkmark & \multicolumn{1}{l}{}      & 15.49       & 34.96       & 49.78      & 13.54      & 41.42      & 58.36     & 35.59               \\
11                                              & \multicolumn{1}{l}{}      & \checkmark & \multicolumn{1}{l}{}      & \checkmark & 16.37       & 36.28       & 47.57      & 14.65      & 42.17      & 58.36     & 35.90               \\
12                                              & \multicolumn{1}{l}{}      & \checkmark & \multicolumn{1}{l}{}      & \multicolumn{1}{l}{}      & 16.15       & 36.06       & 48.01      & 13.14      & 42.61      & 59.69     & 35.94               \\
13                                              & \multicolumn{1}{l}{}      & \multicolumn{1}{l}{}      & \checkmark & \checkmark & 16.59       & 37.61       & 51.55      & 13.94      & 42.17      & 58.94     & 36.80               \\
14                                              & \multicolumn{1}{l}{}      & \multicolumn{1}{l}{}      & \checkmark & \multicolumn{1}{l}{}      & 15.49       & 34.96       & 47.35      & 13.36      & 42.61      & 60.18     & 35.66               \\
15                                              & \multicolumn{1}{l}{}      & \multicolumn{1}{l}{}      & \multicolumn{1}{l}{}      & \checkmark & 14.16       & 33.85       & 46.90      & 14.78      & 43.32      & 58.67     & 35.28                                              \\ \hline
\end{tabular}}
\label{LAYERS}
\end{table}

\begin{figure}[t]
\centering
\includegraphics[width=0.5\textwidth]{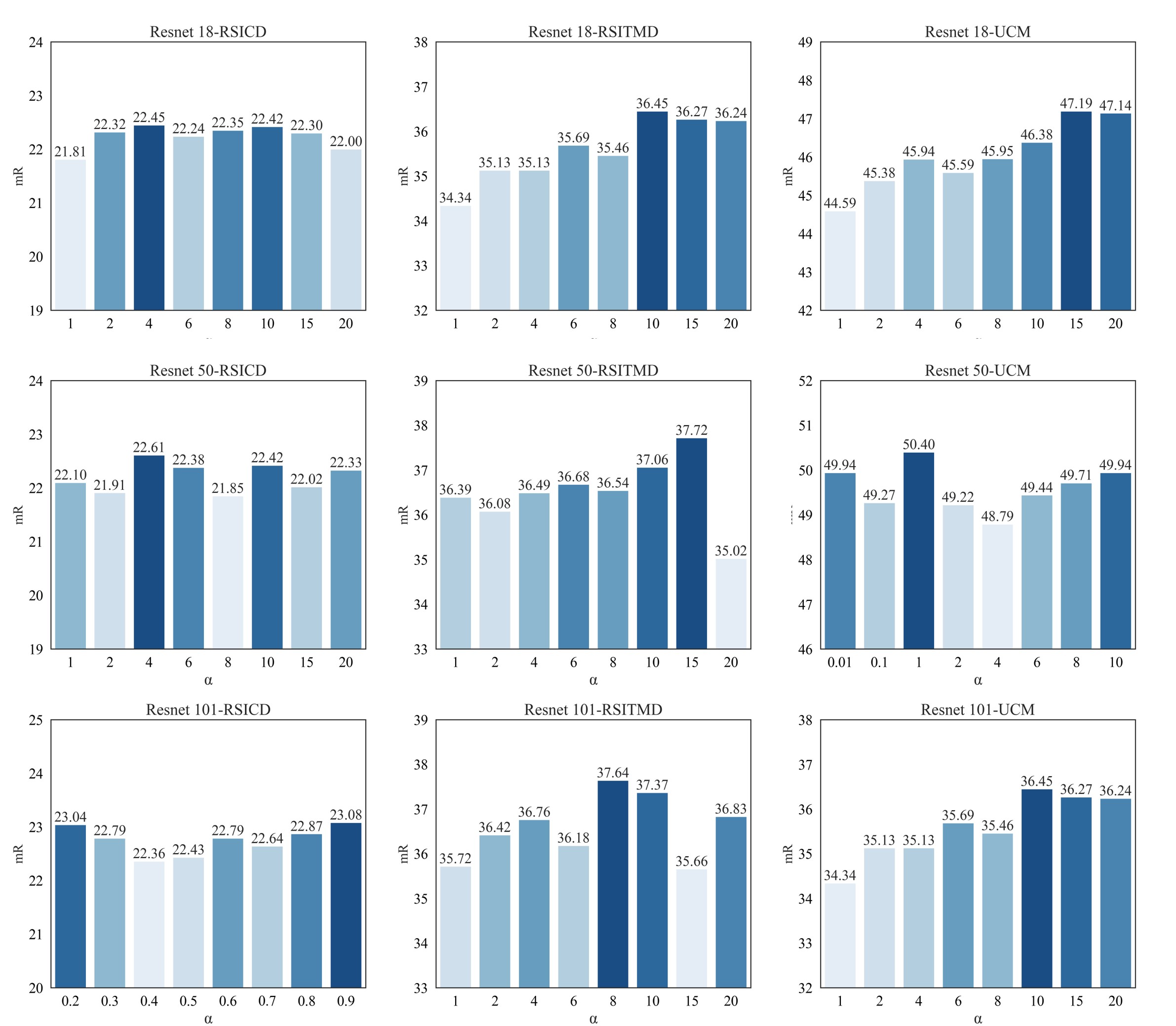} 
 \caption{The parameter search results of $\alpha$ on different datasets and visual backbones.}
\label{alpha}
\end{figure}

\begin{figure}[h]
\centering
\includegraphics[width=0.5\textwidth]{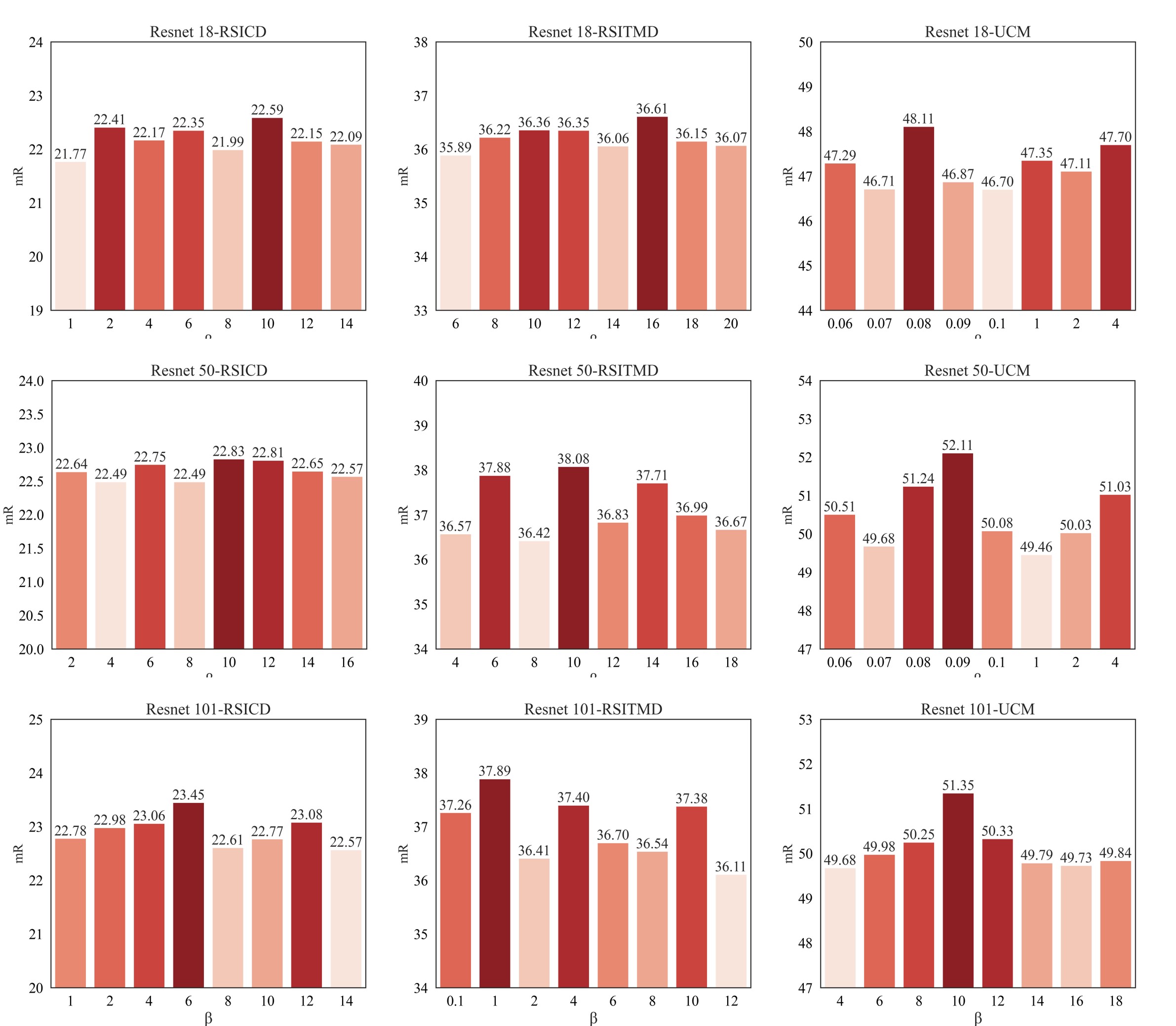} 
 \caption{The parameter search results of $\beta$ on different datasets and visual backbones.}
\label{beta}
\end{figure}

\subsubsection{Parameter Sensitivity Analysis}

We conducted a heuristic parameter search on $\alpha$ and $\beta$ in Equation~\ref{eq8} to explore the model's sensitivity to these parameters.
During the search for each parameter, we conducted experiments in nine scenarios, which corresponded to the combinations of three datasets and three ResNet models.
In each scenario, we initially set $\beta$ to 0 and performed a search for $\alpha$. Then, with optimal $\alpha$ searched fixed, we searched $\beta$. Figure~\ref{alpha} and Figure~\ref{beta} present the search results for $\alpha$ and $\beta$, respectively. To facilitate visualization, each bar chart represents 8 search values that include the optimal results in a given scenario, although our search range extends beyond these 8 values.

From Figure~\ref{alpha}, we observe that when using ResNet18, the optimal $\alpha$ values for RSICD, RSITMD, and UCM are 4, 10, and 15, respectively. When using ResNet50, the optimal $\alpha$ values are 4, 15, and 1. When using ResNet101, these values are 0.9, 8, and 10. The optimal $\alpha$ is sensitive to the size of the visual network model and dataset.

From Figure~\ref{beta}, we can see that when using ResNet18, the optimal $\beta$ values for RSICD, RSITMD, and UCM are 10, 16, and 0.08, respectively. When using ResNet50, the optimal $\beta$ values are 10, 10, and 0.09. When using ResNet101, these values are 6, 1, and 10. The search results for $\beta$ reveal that different networks and datasets have a significant impact on the optimal $\beta$ value. In other words, the optimal $\beta$ is sensitive to the dataset and the visual backbone network used.

\subsection{Other Experiments}

\subsubsection{Exploration on CLIP}

To further validate the generalizability and plug-and-play capability of our MSA, we performed experiments using the robust multimodal pre-trained model, CLIP \cite{radford2021learning}. Our study involved two CLIP-integrated MSA variants—MSA\_CLIP\_RN50 and MSA\_CLIP\_TEXT, as shown in Table~\ref{tab:9}. The MSA\_CLIP\_TEXT variant employs a standard ResNet50 as its visual backbone and utilizes CLIP's text encoder for textual analysis. Conversely, MSA\_CLIP\_RN50 incorporates our MSA approach into the official "CLIP-RN50" model, leveraging its transformer-based text encoder and ResNet-based image encoder for multi-scale image-text feature alignment. The term "CLIP\_RN50 (Fine Tune)" refers to the outcomes of comprehensive fine-tuning on the "CLIP (RN50)" model without any modifications. Additionally, MSA\_18, MSA\_50, and MSA\_101 refer to the configurations from the original manuscript, utilizing ResNet18, ResNet50, and ResNet101 as visual backbones, respectively, paired with BERT for textual analysis.

From Table~\ref{tab:9}, we see that MSA\_CLIP\_TEXT achieved an mR of 37.23, which did not surpass the performance of the three setups using BERT as the textual backbone. This outcome highlights that while CLIP's strength lies in image-text pair representation, using only its text encoder does not fully transfer this powerful representation. On the other hand, utilizing the full capabilities of CLIP yields significantly strong results, with CLI\_RN50 (Fine Tune) reaching an impressive mR of 40.27. This performance benefits from 400 million pretrained image-text pairs. Adapting our MSA strategy to the pretrained "CLIP (RN50)" model, MSA\_CLIP\_RN50 achieved the highest mR of 41.38, demonstrating the usefulness, generalizability, and plug-and-play nature of our MSA strategy.

\begin{table}[]

\caption{\label{tab:9}Results based on CLIP}
\resizebox{\linewidth}{!}{
\begin{tabular}{cccccccc}
\hline
\multirow{2}{*}{\textit{Method}} & \multicolumn{3}{c}{\textit{Sentence Retrieval}} & \multicolumn{3}{c}{\textit{Image Retrieval}} & \multirow{2}{*}{\textit{mR}} \\
                                 & \textit{R@1}   & \textit{R@5}  & \textit{R@10}  & \textit{R@1}  & \textit{R@5} & \textit{R@10} &                              \\ \hline
MSA\_18                          & 19.25          & 37.61         & 50.44          & 13.63         & 40.8         & 57.92         & 36.61                        \\
MSA\_50                          & 16.59          & 40.04         & 51.33          & 15.53         & 44.20         & 60.8          & 38.08                        \\
MSA\_101                         & 15.93          & 38.50          & 50.88          & 14.96         & 45.22        & 61.86         & 37.89                        \\
MSA\_CLIP\_TEXT                  & 17.92          & 38.05         & 51.77          & 13.94         & 43.58        & 58.14         & 37.23                        \\
CLIP\_RN50 (Fine Tune)           & 18.58          & 41.81         & 51.33          & \textbf{16.42}         & \textbf{47.35}        & \textbf{66.15}         & 40.27                        \\
MSA\_CLIP\_RN50                  & \textbf{22.35}          & \textbf{42.92}         & \textbf{55.75}          & 15.18         & \textbf{47.35 }       &64.73         & \textbf{41.38 }                       \\ \hline
\end{tabular}}
\end{table}

To further explore the generalizability of MSA, we conducted detailed ablation studies on the MSA\_CLIP\_TEXT configuration to assess the contribution of each component of MSA when using CLIP's text encoder as the textual feature extractor. These experiments on the RSITMD dataset involved variations: BASE, which uses ResNet50 as the visual backbone and CLIP’s text encoder as the textual backbone; BASE+M, which adds traditional multi-scale fusion strategies; BASE+M+A+B, which further includes our multi-scale cross-modal alignment transformer and multi-scale cross-modal semantic alignment loss; and BASE+M+A+B+C, which incorporates our proposed cross-scale multimodal semantic consistency loss. Results are shown in Table~\ref{tab:10}.

\begin{table}[]

\caption{\label{tab:10}Ablation study using CLIP's text encoder as the feature extractor}
\resizebox{\linewidth}{!}{
\begin{tabular}{cccccccc}
\hline
\multirow{2}{*}{\textit{Method}} & \multicolumn{3}{c}{\textit{Sentence Retrieval}} & \multicolumn{3}{c}{\textit{Image Retrieval}} & \multirow{2}{*}{\textit{mR}} \\
                                 & \textit{R@1}   & \textit{R@5}  & \textit{R@10}  & \textit{R@1}  & \textit{R@5} & \textit{R@10} &                              \\ \hline
BASE                             & 15.71          & 31.64         & 46.24          & 11.95         & 41.77        & \textbf{60.27}         & 34.59                        \\
BASE+M                           & 13.50          & 30.09         & 44.91          & 11.15         & 38.32        & 54.03         & 32.00                        \\
BASE+M+A+B                       & 15.49          & 36.06         & 51.55          & 13.36         & 40.84        & 60.18         & 36.25               \\
BASE+M+A+B+C (MSA)               & \textbf{17.92}          & \textbf{38.05}         & \textbf{51.77}          & \textbf{13.94}         & \textbf{43.58}       & 58.14         & \textbf{37.23}               \\ \hline
\end{tabular}}
\end{table}

The results demonstrate that BASE+M did not improve retrieval performance compared to BASE, consistent with findings using BERT as the text encoder in the original manuscript. BASE+M+A+B showed a significant improvement over BASE and BASE+M, with increases of 1.66 and 4.25 in mR respectively, confirming the effectiveness of the proposed image-text alignment at various scales. Finally, BASE+M+A+B+C (MSA) further increased the mR by 0.98 compared to BASE+M+A+B, validating the efficacy of the proposed cross-scale multimodal semantic consistency loss.

These experimental results affirm that the proposed MSA strategy is effective across different visual backbones (ResNet18, 50, 101) and textual backbones (BERT and CLIP's text encoder), further demonstrating MSA's strong generalizability and plug-and-play capability.

\begin{figure}[t]
\centering
\includegraphics[width=1\linewidth]{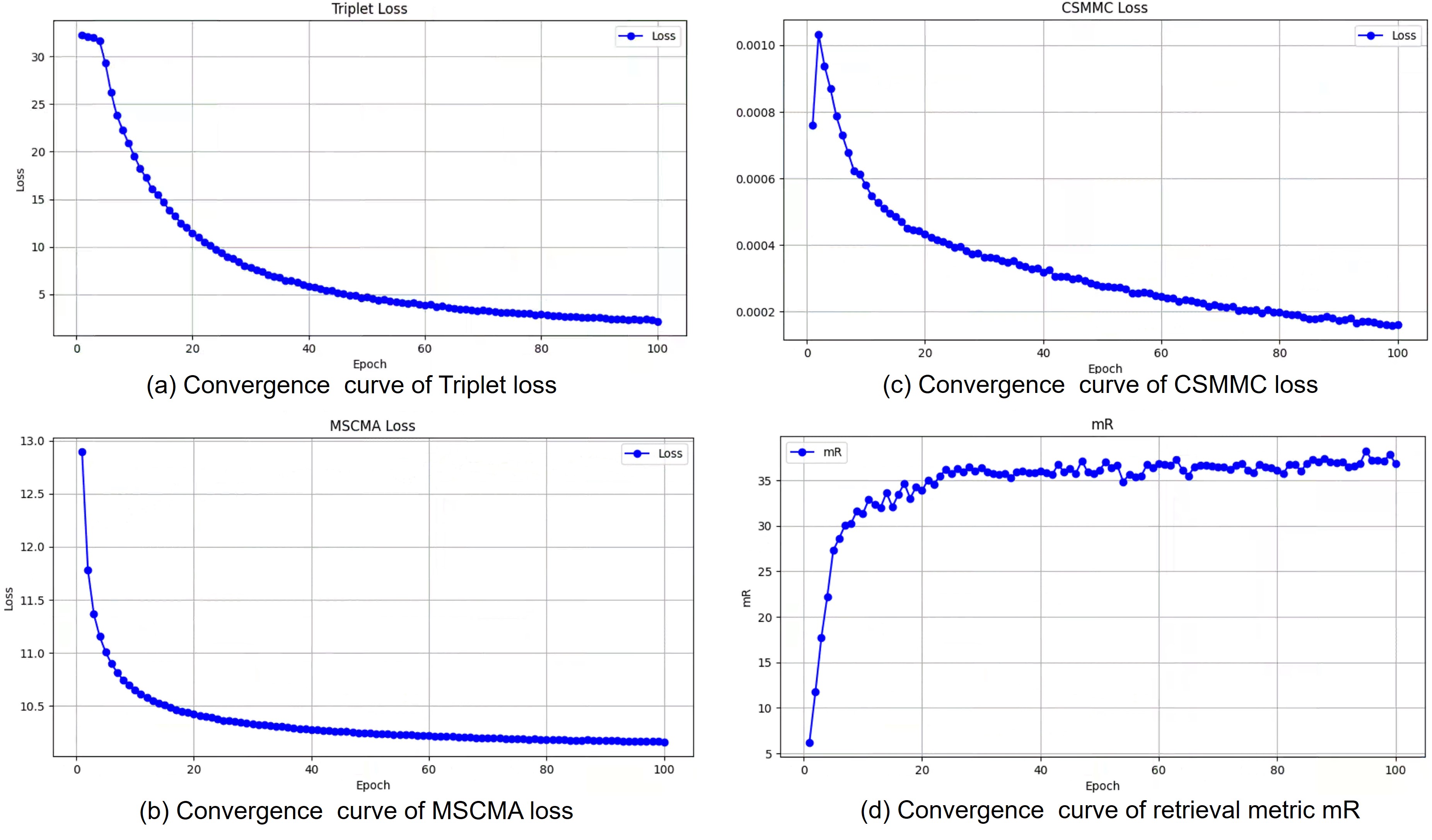} 
 \caption{Convergence Analysis of MSA}
\label{Convergence}
\end{figure}

\begin{figure}[t]
\centering
\includegraphics[width=0.5\textwidth]{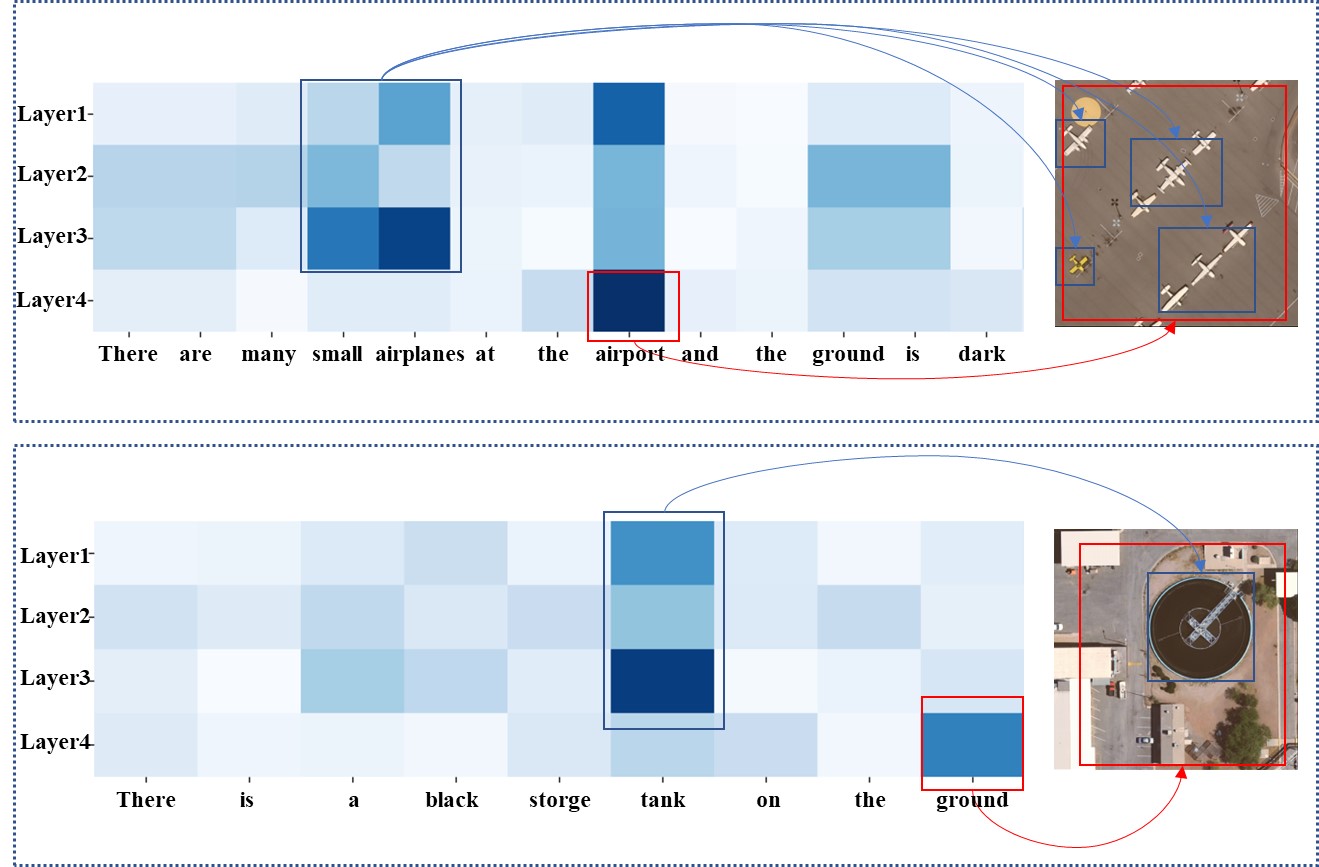} 
 \caption{Visualization of cross-attention in MSCMAT. The horizontal axis represents the words in the textual description, while the vertical axis represents the different layers of image features from ResNet50. The darker the color, the higher the value of cross-attention.}
\label{att_keshihua}
\end{figure}

\subsubsection{Comparison with Other Multi-scale Methods}

Currently, several RSITR methods such as AMFMN \cite{yuan2021exploring}, FAAMI \cite{zheng2023fine}, GaLR \cite{yuan2022remote}, MSITA \cite{chen2023multiscale}, and SWAN \cite{pan2023reducing} consider multi-scale nature of remote sening images. These methods leverage the multi-scale nature of remote sensing images and achieve commendable retrieval results. We have conducted a detailed comparison of our method with these multi-scale approaches, as shown in Table~\ref{tab:0}. Table~\ref{tab:0} illustrates that the key distinction of MSA compared to other methods is its focus on image-text alignment across different scales. Existing approaches effectively utilize multi-scale image features through the design of multi-scale image fusion modules. Building upon these techniques, our study introduces image-text alignment at various scales, further enhancing the utilization of multi-scale features and achieving improved retrieval performance.

To evaluate the impact of incorporating multi-scale image-text alignment, we compared our method with the AMFMN approach. AMFMN utilizes ResNet18 as an image feature extractor and introduces the MVSA module, which integrates multi-scale image features. For text processing, AMFMN employs a bidirectional GRU as its text encoder. To ensure a fair comparison, we modified AMFMN by replacing its text encoder with the same BERT model used in our MSA approach, while retaining the ResNet18 and MVSA module for image processing. This modified version is referred to as AMFMN\*. The key difference between AMFMN* and our MSA\_18 is that MSA\_18 implements multi-scale image-text alignment. Our comparative analysis between MSA\_18 and AMFMN\* is presented in Table~\ref{tab:11}.Table~\ref{tab:11} shows that using the same backbone, MSA\_18 achieved a higher mR value of 36.61, compared to 33.71 for AMFMN*. This demonstrates the effectiveness of our proposed multi-scale image-text alignment and underscores the superiority of the MSA method presented in our paper.

\begin{table}[]

\caption{\label{tab:11}Detailed comparison between MSA and AMFMN on RSITMD}
\resizebox{\linewidth}{!}{
\begin{tabular}{cccccc}
\hline
Method  & Multi-scale fusion & Alignment at different scales & Image backbone & Text backbone & mR \\ \hline
AMFMN   & Yes                & No                            & ResNet18       & GRU           & 29.72        \\
AMFMN*  & Yes                & No                            & ResNet18       & BERT          & 33.71        \\
MSA\_18 & Yes                & Yes                           & ResNet18       & BERT          & 36.61        \\ \hline
\end{tabular}}
\end{table}

\subsubsection{Convergence Analysis}
We verified the convergence of MSA\_50 on the RSITMD dataset by documenting the average loss per epoch for the three loss functions used in this study—Triplet loss, MSCMA loss, and CSMMC loss—over 100 epochs, as shown in Figure~\ref{Convergence} (a), (b), and (c). It is important to note that the recorded values are the raw loss values, unadjusted by the combination coefficients of 1, 15, and 10, respectively, for each loss function. Additionally, we tracked the retrieval metric mR for each epoch, as displayed in Figure~\ref{Convergence} (d).

From Figure~\ref{Convergence} (a) to (c), it is evident that each of our loss functions—both the primary losses, Triplet loss and MSCMA loss, and the auxiliary loss, CSMMC loss—decreases over the training process. The decline becomes very slight after 60 epochs and stabilizes around the 100th epoch. Additionally, observing Figure~\ref{Convergence} (d), we notice that the retrieval metric mR consistently increases throughout the training and gradually stabilizes with minimal changes after 60 epochs. Based on these observations, we can conclude that our algorithm demonstrates good convergence.

\begin{figure*}[t]
\centering
\includegraphics[width=1\textwidth]{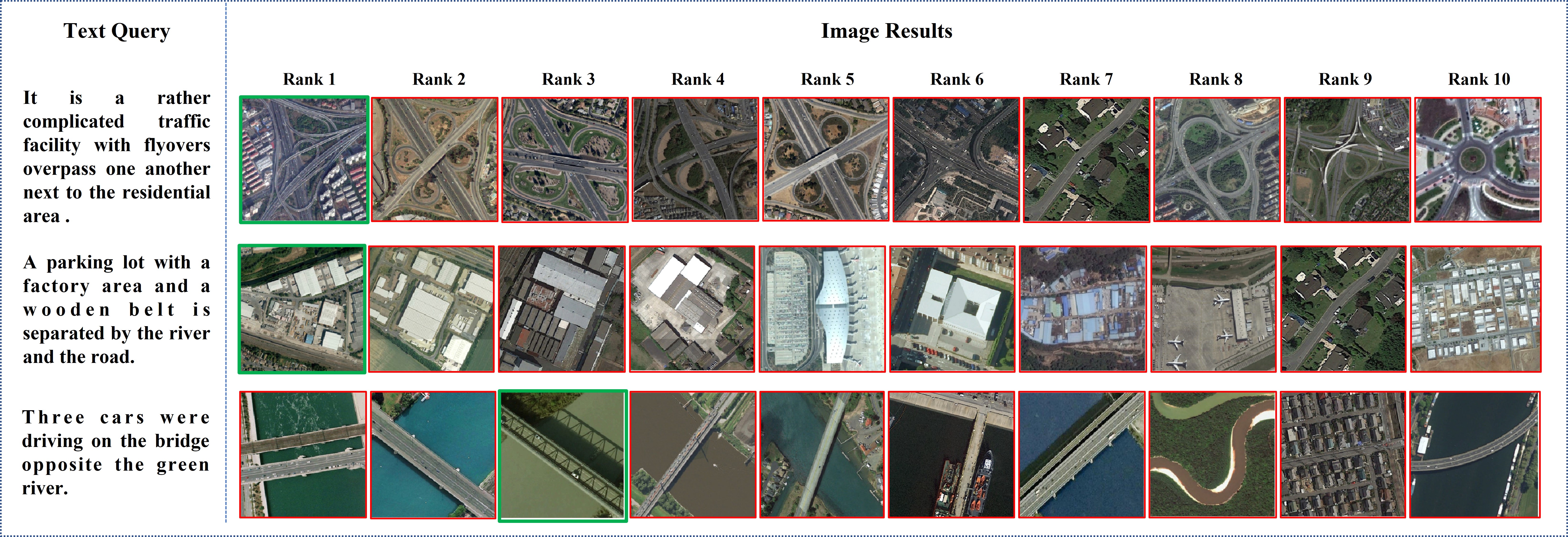} 
 \caption{Presentation of retrieval results.}
\label{results_keshihua}
\end{figure*}

\subsubsection{Runtime Analysis}

Based on the methodology described in the literature \cite{yuan2021exploring}, we measured the feature extraction time for a single image-text pair on the same computer to assess the runtime of our MSA method compared to other approaches. The results, displayed in Table~\ref{tab:12}, show that MSA-101, MSA-50, and MSA-18 have runtimes of 34 ms, 22 ms, and 16 ms respectively, which are advantageous compared to most other algorithms. This efficiency is due to the MSA's alignment operations being performed during training, resulting in a simpler structure—consisting only of a text encoder and an image encoder—during testing. This leads to lower inference times. Notably, MSA-18's runtime is only 1 ms longer than the lightweight RSITR method, LW-MCR, yet MSA-18 surpasses LW-MCR in retrieval performance on the RSITMD, RSICD, and UCM Captioning datasets, demonstrating both the effectiveness and efficiency of our method.

 \begin{table}[]

\caption{\label{tab:12}Comparison of runtime}
\resizebox{\linewidth}{!}{
\begin{tabular}{ccccccccc}
\hline
Method    & VSE++      & MTFN  & MCRN & SWAN  & LW-MCR     & AMFMN    & GALR    & CABIR   \\
Time (ms) & 21         & 22    & 34   & 27    & 15         & 60       & 71      & 155     \\ \hline
Method    & HpyerMatch & SSJDN & MGRM & MSITA & VIT + BERT & MSA-101 & MSA-50 & MSA-18 \\
Time (ms) & 37         & 39    & 125  & 50    & 25         & 34       & 22      & 16      \\ \hline
\end{tabular}}
\end{table}

\subsubsection{Visualization of Cross-Attention in MSCMAT}

To understand how the model attends to relevant information during the alignment process, we visualized the cross-attention mechanism in MSCMAT. This visualization helps us gain insights into the alignment between image and text at different scales. 
Here are the visualization results for two test samples, as shown in Figure~\ref{att_keshihua}.

The first sample in Figure~\ref{att_keshihua} illustrates the cross-attention values between an RS image about an airport and the text "There are many small airplanes at the airport and the ground is dark" at different scales. We can observe that the image features from Layer 4 of ResNet50 (large scale) have the highest cross-attention value with the word "airport," while the image features from the other three layers (smaller scales) focus more on "small airplanes."

The second sample displays the cross-attention values between an RS image of a storage tank and the text "There is a black storage tank on the ground" at different scales. We can see that the image features from Layer 4 of ResNet50 (largest scale) have the highest cross-attention value with the word "ground," while the image features from the other three layers (smaller scales) pay more attention to "tank."

The phenomena depicted in  Figure~\ref{att_keshihua} demonstrate that our model can adaptively learn the alignment between image and text at different scales. Larger image scales correspond to "larger-scale" textual descriptions, while smaller image scales correspond to more detailed aspects. This qualitative observation highlights the effectiveness of the proposed method.

\subsubsection{Visualization of Retrieval Results}

To visually demonstrate the retrieval performance of the proposed method, we conducted visualizations of text-to-image retrieval on a subset of the test set from RSITMD.

The experimental results, as shown in Figure~\ref{results_keshihua}, display the retrieval outcomes. The first retrieval query in the image illustrates a text query about an interchange bridge. We can observe that the top ten retrieved RS images all depict interchange bridges, with the ground truth image ranking first. The description in this sentence is highly detailed, mentioning "residential area" and "flyovers overpass," among others. Our method effectively captures these different objects, resulting in excellent retrieval performance.

The second retrieval outcome accurately places the ground truth image in Rank 1. The model also comprehends the complex semantic description of "separated by the river and the road." Additionally, the model successfully aligns objects of different scales mentioned in the sentence, such as "factory area," "wooden belt," "river," and "road." This further demonstrates our method's capability to capture multi-scale alignment.

In the third retrieval outcome, the ground truth image ranks third. Although it is not in the top 1 position, the first five retrieved results all depict "bridges with cars on them". This indicates that our method can capture information about objects at different scales and perform multi-scale alignment. However, this also reveals some limitations of our method regarding the understanding of the quantity of "cars." In the first two retrieved images, there are more than three cars on the bridge (please zoom in to see). This might be due to insufficient textual descriptions regarding the quantity in the dataset, leading to limitations in the model's understanding of more detailed descriptions. We will address this issue in future work.

\section{Conclusion}
Existing multi-scale RSITR methods predominantly focus on aligning multi-scale fused image features with text without considering distinct scale-wise alignments. Our proposed Multi-Scale Alignment (MSA) method addresses this gap by introducing (1) Multi-scale Cross-Modal Alignment Transformer (MSCMAT), (2) multi-scale cross-modal semantic alignment loss, and (3) cross-scale multi-modal semantic consistency loss. The MSCMAT effectively calculates cross-attention between image features at individual scales and corresponding local text features, utilizing global textual information to create a context-rich matching score matrix. Our proposed two losses ensure robust semantic alignment in single-scale and coherence across all scales. Comprehensive evaluations on multiple datasets confirm the effectiveness of MSA, showing significant improvements over state-of-the-art methods. While our method advances RSITR, it has limitations such as not testing on a wider range of visual backbones and limited capture of target quantity information. Future efforts will focus on addressing these areas to further refine our approach.

\section*{Acknowledgments}
This work was supported by the National Natural Science Foundation of China under Grant 62271377, 62201407, 62171347, the National Key R\&D Program of China under Grant No. 2021ZD0110400, the Key Research and Development Program of Shannxi (Program No. 2021ZDLGY01\-06 and No. 2022ZDLGY01\-12), and the Key Scientific Technological Innovation Research Project by Ministry of Education.



\bibliographystyle{IEEEtran} 
\bibliography{ref}

\end{document}